\documentclass{article}
\usepackage{log_2025}				

\usepackage{booktabs}						
\usepackage{multirow}						
\usepackage{amsfonts}						
\usepackage{graphicx}						
\usepackage{duckuments}						

\usepackage[
     backend=biber,
     style=numeric-comp,
     backref=true,
     natbib=true]{biblatex}
\author[Sverdlov et al.]{%
Yonatan Sverdlov\\
Technion – Israel Institute of Technology\\
\email{yonatans@campus.technion.ac.il}
\And
Yair Davidson\\
Technion – Israel Institute of Technology\\
\email{yairdavidson@campus.technion.ac.il}
\And
Nadav Dym\\
Technion – Israel Institute of Technology\\
\email{nadavdym@technion.ac.il}
\And
Tal Amir\\
Technion – Israel Institute of Technology\\
\email{talamir@technion.ac.il}
}

\newcommand\TheDocumentTitle{FSW-GNN: A Bi-Lipschitz WL-Equivalent Graph Neural Network}
\title{\TheDocumentTitle}

\newcommand\PrintTheBibliography{\printbibliography}

\input{base/packages.tex}
\makeatletter
\def\input@path{{csv/}{text/}{tables/}{figures/}}
\makeatother

\providetoggle{show_our_comments}
\settoggle{show_our_comments}{false} 

\newcommand{\nd}[1]{
\iftoggle{show_our_comments}{{\color{red}{\bf[Nadav:} \textup{#1}{\bf]}}}{}}

\newcommand{\ta}[1]{\iftoggle{show_our_comments}{{\color{blue}{\bf[Tal:} \textup{#1}{\bf]}}}{}}

\newcommand{\ys}[1]{\iftoggle{show_our_comments}{{\color{green}{\bf[Yonatan:} \textup{#1}{\bf]}}}{}}

\newif\ifinappendix
\inappendixfalse

\let\oldappendix\appendix
\renewcommand{\appendix}{%
  \oldappendix%
  \inappendixtrue%
}

\newcommand{\onmain}[1]{\ifinappendix\else#1\fi}

\graphicspath{{./figures/}}

\makeatletter
\@ifpackageloaded{amsthm}{}
{
    \ifcsdef{proof}{\let\proof\relax}{}
    \ifcsdef{endproof}{}{}
}
\makeatother

\usepackage{amsmath}
\usepackage{amsthm}
\usepackage{amsfonts}
\usepackage{amssymb}
\usepackage{mathtools}
\usepackage{thmtools}
\usepackage{bm}           
\usepackage{booktabs} 
\usepackage{amsmath}  

\usepackage{pgfplots}

\makeatletter
\def\cleartheorem#1{%
    \expandafter\let\csname#1\endcsname\relax
    \expandafter\let\csname c@#1\endcsname\relax
}
\makeatother

\makeatletter
\def\cleartheorems#1{ \@for\tname:=#1\do{\cleartheorem\tname} }
\makeatother

\NewDocumentCommand \newtheoremHelperMacro{ m m o m o }
{       
    \IfBooleanTF{#1} 
    {        
        \IfValueTF{#3}
        {
            \IfValueTF{#5}
            {
                \newtheorem*{#2}[#3]{#4}[#5]
            }
            {
                \newtheorem*{#2}[#3]{#4}
            }
        }
        {
            \IfValueTF{#5} 
            {
                \newtheorem*{#2}{#4}[#5]
            }
            {
                \newtheorem*{#2}{#4}
            }
        }
    }
    {        
        \IfValueTF{#3}
        {
            \IfValueTF{#5}
            {
                \newtheorem{#2}[#3]{#4}[#5]
            }
            {
                \newtheorem{#2}[#3]{#4}
            }
        }
        {
            \IfValueTF{#5} 
            {
                \newtheorem{#2}{#4}[#5]
            }
            {
                \newtheorem{#2}{#4}
            }
        }
    } 
}

\NewDocumentCommand \trynewtheorem{ s m o m o }
{
    \ifcsundef{#2}{\newtheoremHelperMacro{#1}{#2}[#3]{#4}[#5]}{}
}

\NewDocumentCommand \forcenewtheorem{ s m o m o }
{
    \ifcsdef{#2}{\cleartheorem{#2}}{}
    \newtheoremHelperMacro{#1}{#2}[#3]{#4}[#5]
}

\theoremstyle{plain} 

\forcenewtheorem{theorem}{Theorem}[section]
\forcenewtheorem{lemma}[theorem]{Lemma}
\forcenewtheorem{claim}[theorem]{Claim}
\forcenewtheorem{proposition}[theorem]{Proposition}
\forcenewtheorem{corollary}[theorem]{Corollary}

\forcenewtheorem*{theorem*}{Theorem}
\forcenewtheorem*{lemma*}{Lemma}
\forcenewtheorem*{claim*}{Claim}
\forcenewtheorem*{proposition*}{Proposition}
\forcenewtheorem*{corollary*}{Corollary}

\theoremstyle{definition} 

\forcenewtheorem{definition}[theorem]{Definition}
\forcenewtheorem{remark}[theorem]{Remark}
\forcenewtheorem{note}[theorem]{Note}
\forcenewtheorem{question}{Question}
\forcenewtheorem{example}[theorem]{Example}

\forcenewtheorem*{definition*}{Definition}
\forcenewtheorem*{remark*}{Remark}
\forcenewtheorem*{note*}{Note}
\forcenewtheorem*{question*}{Question}
\forcenewtheorem*{example*}{Example}

\newcommand{\of}[1]{{\left({#1}\right)}} 

\newcommand{\setstspace}{{\ \,}}
\newcommand{\setst}[2]{ { \left\{ {#1} \setstspace\middle|\setstspace {#2} \right\} } }

\newcommand{\br}[1]{{\left({#1}\right)}} 
\newcommand{\brs}[1]{{\left[{#1}\right]}} 
\newcommand{\brc}[1]{{\left\{{#1}\right\}}} 
\newcommand{\bra}[1]{{\left\langle{#1}\right\rangle}} 
\newcommand{\norm}[1]{{\left\lVert{#1}\right\rVert}} 
\newcommand{\abs}[1]{{\left\lvert{#1}\right\rvert}} 

\DeclareDocumentCommand \expect{ o m } {{ \IfNoValueTF{#1}{\mathbb{E}\brs{#2}}{\mathbb{E}_{#1}\brs{#2}} }}
\DeclareDocumentCommand \stdev{ o m } {{ \IfNoValueTF{#1}{\textup{STD}\brs{#2}}{\textup{STD}_{#1}\brs{#2}} }}
\DeclareDocumentCommand \variance{ o m } {{ \IfNoValueTF{#1}{\textup{Var}\brs{#2}}{\textup{Var}_{#1}\brs{#2}} }}

\DeclareDocumentCommand \rootp{ o m } {{ \IfNoValueTF{#1} { {\sqrt[\leftroot{-2}\uproot{2}{p}]{#2}} } { {\sqrt[\leftroot{-2}\uproot{2}{#1}]{#2}} } }}

\newcommand{\eqdef}{\coloneqq} 

\newcommand{\eqxspace}{{\hspace{0.0pt}}} 
\newcommand{\eqx}[1][]{{\eqxspace\overset{{\textup{#1}}}{=}\eqxspace}}
\newcommand{\leqx}[1][]{{\eqxspace\overset{{\textup{#1}}}{\leq}\eqxspace}}

\newcommand{\eqdefx}[1][]{{\eqxspace\overset{{\textup{#1}}}{\eqdef}\eqxspace}}

\newcommand{\argdot}{{\hspace{0.18em}{{\bm{\cdot}}}\hspace{0.18em}}} 

\DeclareSymbolFont{tipa}{T3}{cmr}{m}{n}
\DeclareMathAccent{\invbreve}{\mathalpha}{tipa}{16}

\usepackage{scalerel,stackengine}
\stackMath
\newcommand\vwidehat[1]{%
\savestack{\tmpbox}{\stretchto{%
  \scaleto{%
    \scalerel*[\widthof{\ensuremath{#1}}]{\kern-.6pt\bigwedge\kern-.6pt}%
    {\rule[-\textheight/2]{1ex}{\textheight}}
  }{\textheight}%
}{0.5ex}}%
\stackon[1pt]{#1}{\tmpbox}%
}

\def\eqref#1{equation~\ref{#1}}

\def\1{\bm{1}}

\DeclareMathAlphabet{\mathsfit}{\encodingdefault}{\sfdefault}{m}{sl}
\SetMathAlphabet{\mathsfit}{bold}{\encodingdefault}{\sfdefault}{bx}{n}

\newcommand{\R}{\mathbb{R}}

\newcommand{\wleq}{{\mathrel{\overset{\tiny\textup{WL}}{\sim}}}}

\newcommand{\rhods}{{\rho_{\textup{DS}}}}

\DeclareDocumentCommand \TMD{ o } 
{ {\textup{TMD}\IfNoValueF{#1}{^{{#1}}}{}} }

\newcommand{\X}{\textbf{X}}
\newcommand{\Y}{\textbf{Y}}
\newcommand{\Z}{\textbf{Z}}
\newcommand{\C}{\textbf{C}}

\renewcommand{\upsilon}{\eta} 

\newcommand{\GNof}[1]{{\G_{\leq N}\of{{#1}}}}
\newcommand{\GNofe}[1]{{\G_{= N}\of{{#1}}}}

\newcommand{\GNd}{{\GNof{\RR^d}}}

\newcommand{\G}{\mathcal{G}}

\newcommand{\V}{\mathcal{V}}

\newcommand{\D}{\mathcal{D}}
\newcommand{\N}{\mathcal{N}}
\renewcommand{\P}{\mathcal{P}}

\newcommand{\HH}{\mathcal{H}}

\newcommand{\A}{\textbf{A}}
\newcommand{\B}{\textbf{B}}
\renewcommand{\S}{\mathcal{S}}
\newcommand{\NN}{\mathbb{N}}

\newcommand{\RR}{\mathbb{R}}

\newcommand{\sort}{\textup{sort}}

\DeclareDocumentCommand \bx{ o } {{ {\boldsymbol{x}} \IfNoValueF {#1} {^{\br{#1}}} }}

\DeclareDocumentCommand \btx{ o } {{ {\tilde{\boldsymbol{x}}} \IfNoValueF {#1} {^{\br{#1}}} }}

\DeclareDocumentCommand \by{ o } {{ {\boldsymbol{y}} \IfNoValueF {#1} {^{\br{#1}}} }}

\DeclareDocumentCommand \bv{ o } {{ {\boldsymbol{v}} \IfNoValueF {#1} {^{\br{#1}}} }}
\DeclareDocumentCommand \xik{ o } {{ {\xi} \IfNoValueF {#1} {^{\br{#1}}} }}

\DeclareDocumentCommand \bu{ o } {{ {\boldsymbol{u}} \IfNoValueF {#1} {^{\br{#1}}} }}

\DeclareDocumentCommand \Wass{ o o } {{ {\mathcal{W}} \IfNoValueF {#1} {_{{#1}}} \IfNoValueF {#2} {^{{#2}}} }}

\DeclareDocumentCommand \bp{ o } {{ {\boldsymbol{p}} \IfNoValueF {#1} {^{\br{#1}}} }}
\DeclareDocumentCommand \bq{ o } {{ {\boldsymbol{q}} \IfNoValueF {#1} {^{\br{#1}}} }}
\DeclareDocumentCommand \bw{ o } {{ {\boldsymbol{w}} \IfNoValueF {#1} {^{\br{#1}}} }}
\DeclareDocumentCommand \btw{ o } {{ {\tilde{\boldsymbol{w}}} \IfNoValueF {#1} {^{\br{#1}}} }}

\newcommand{\bmsg}{\boldsymbol{m}}
\newcommand{\bh}{\boldsymbol{h}}

\newcommand{\bX}{\boldsymbol{X}}
\newcommand{\btX}{\tilde{\boldsymbol{X}}}

\newcommand{\bz}{\boldsymbol{z}}

\newcommand{\bzero}{\boldsymbol{0}}

\newcommand{\emb}{{{E}}}
\newcommand{\embfsw}{{{E}_{\textup{FSW}}}}
\newcommand{\embfswglob}{{{E}_{\textup{FSW}}^{\textup{Glob}}}}
\DeclareDocumentCommand \embfswt{ o } {{ {{E}}_{\textup{FSW}} \IfNoValueTF {#1} {^\br{t}} {^\br{{#1}}} }}
\DeclareDocumentCommand \embfswm{ o } {{ {{E}}^{\textup{FSW}} \IfNoValueTF {#1} {_m} {_{{#1}}} }}
\DeclareDocumentCommand \embt{ o } {{ {\textup{FSW}}\IfNoValueTF {#1} {^\br{t}} {^\br{{#1}}} }}

\DeclareDocumentCommand \Lp{ o } {{ \IfNoValueTF{#1} {{L}^p}{{L}^{#1}} }}
\DeclareDocumentCommand \LpR{ o } {{ \IfNoValueTF{#1} {{L}^p\of{\RR}}{{L}^{#1}\of{\RR}} }}

\newcommand{\tmd}{\mathrm{TMD}}
\newcommand{\TD}{\mathrm{TD}}
\newcommand{\tree}{\mathbf{T}}
\newcommand{\blankT}{\Bar{\tree}_0}

\begin{document}
%
%

\maketitle 
\begin{abstract} 
%
%

Famously, the ability of Message Passing Neural Networks (MPNN) to distinguish between graphs is limited to graphs separable by the Weisfeiler-Lemann (WL) graph isomorphism test, and the strongest MPNNs, in terms of separation power, are WL-equivalent. 
However, it was demonstrated that the quality of separation provided by standard WL-equivalent MPNN can be very low, resulting in WL-separable graphs being mapped to very similar, hardly distinguishable outputs. This phenomenon can be explained by the recent observation that standard MPNNs are not lower-Lipschitz. 
This paper addresses this issue by introducing FSW-GNN, the first MPNN that is fully bi-Lipschitz with respect to standard WL-equivalent graph metrics.   Empirically, we show that our MPNN is competitive with standard MPNNs for several graph learning tasks and is far more accurate in long-range tasks, due to its ability to avoid oversmoothing and oversquashing. Our code is available at \url{https://github.com/yonatansverdlov/Over-squashing}.
\end{abstract}

\ta{Hide the TOC by setting \texttt{show\_our\_comments} to false in \texttt{base/settings.tex}.}

\iftoggle{show_our_comments}{\tableofcontents}{}

\iftoggle{show_our_comments}{\newpage
\section*{General Comments}
\begingroup
\color{blue}

This section disappears when you disable \texttt{show\_our\_comments} in \texttt{base/settings.tex}.

\endgroup

}{}

\section{Introduction}
\label{sec_introduction}
Graph neural networks are a central research topic in contemporary machine learning. As shown by \citet{Gilmer2017NeuralMP}, many of the most popular models can be seen as instantiations of Message Passing Neural Networks (MPNNs). 

A well-known limitation of MPNNs is that they cannot differentiate between all distinct pairs of graphs. In fact, a pair of distinct graphs that cannot be separated by the Weisfeiler-Lehman (WL) graph isomorphism test will not be separated by \emph{any} MPNN \citep{xu2018how}. Accordingly, the most expressive MPNNs are those that are  \emph{WL-equivalent}, which means they can separate all pairs of graphs that are separable by WL. WL-equivalent MPNNs were proposed in the seminal works of \citet{xu2018how,morris2019WL}, and the complexity of these constructions was later improved in \citep{aamand2022exponentially,amir2023neural}. 

While separation is theoretically guaranteed with WL-equivalent MPNNs, in some cases, their separation in practice is so weak that it cannot be observed with 32-bit floating-point number; see \cite{pmlr-v235-bravo24a}. Moreover, using many MPNN iterations often leads to almost-identical node features (oversmoothing), or features that are barely affected by changes in far-off nodes (oversquashing). These observations motivate the development of \emph{quantitative} estimates of MPNN separation by means of bi-Lipschitz stability guarantees. These guarantees would ensure that  Euclidean distances in the MPNN feature space are neither much larger nor much smaller than distances in the original graph space, which are defined by a suitable metric on graphs. Consequently, as we shall see, detrimental phenomena like oversmoothing and oversquashing can be fundamentally mitigated.

\begin{figure}[t]
\centering
\begin{minipage}{0.7\textwidth}
    \definecolor{darkgreen}{rgb}{0.0, 0.5, 0.0} 
\definecolor{brown}{rgb}{0.6, 0.4, 0.2} 
\begin{tikzpicture}
\begin{axis}[
    width=\linewidth, 
    height=0.65\linewidth, 
    xlabel={Problem radius},
    ylabel={Accuracy},
    xlabel style={font=\small},
    ylabel style={font=\small},
    legend style={
        at={(0.04,0.04)}, 
        anchor=south west, 
        font=\small,
        nodes={scale=1, transform shape}, 
        legend columns=1 
    },
    grid=both,
    grid style={opacity=0.5},
    tick label style={
        /pgf/number format/fixed, 
        /pgf/number format/precision=2, 
        /pgf/number format/zerofill=false, 
    },
    yticklabel={\pgfmathprintnumber{\tick}\kern0.1em\%},
    scaled ticks=false 
        ]
\addplot[
    color=orange, 
    mark=square*, 
    mark size=2pt,
    line width=1pt,
    solid]
    table[x=Number Of MPNN Layers, y=FSW-GNN (Orange), col sep=comma] {results_trees.csv};
\addlegendentry{FSW-GNN}
\addplot[
    color=brown, 
    mark=square*, 
    mark size=2pt,
    line width=1pt,
    dash dot]
    table[x=Number Of MPNN Layers, y=Sort-MPNN (Brown), col sep=comma] {results_trees.csv};
\addlegendentry{Sort-MPNN}
\addplot[
    color=purple, 
    mark=diamond*, 
    mark size=3pt,
    line width=1pt,
    dotted]
    table[x=Number Of MPNN Layers, y=GAT (Purple), col sep=comma] {results_trees.csv};
\addlegendentry{GAT}
\addplot[
    color=red, 
    mark=triangle*, 
    mark size=3pt,
    line width=1pt,
    dashed]
    table[x=Number Of MPNN Layers, y=GIN (Blue), col sep=comma] {results_trees.csv};
\addlegendentry{GIN}
\addplot[
    color=blue, 
    mark=*, 
    mark size=2pt,
    line width=1pt,
    solid] 
    table[x=Number Of MPNN Layers, y=GCN (Green), col sep=comma] {results_trees.csv};
\addlegendentry{GCN}
\addplot[
    color=cyan, 
    mark=star, 
    mark size=3pt,
    line width=1pt,
    dashed]
    table[x=Number Of MPNN Layers, y=GGNN (Red), col sep=comma] {results_trees.csv};
\addlegendentry{GGNN}
\end{axis}
\end{tikzpicture}
    \end{minipage}
    \caption{FSW-GNN handles tasks with large radius better than standard MPNNs, which are more prone to oversmoothing and oversquashing. Task taken from \cite{alon}}
    \label{fig:Trees}

\end{figure}


This paper introduces a novel MPNN, called \emph{FSW-GNN} (\emph{Fourier Sliced-Wasserstein GNN}), which is  bi-Lipschitz  with respect to two WL-equivalent graph metrics: (a) the Doubly Stochastic (DS) metric of \cite{grohe}, and (b) the Tree Mover's Distance
(TMD) metric of \citet{TMD}. 
Empirically, we show that FSW-GNN performs comparably or better than prevalent MPNNs on standard learning tasks, and  achieves significantly superior performance in long-range tasks, i.e., tasks that require a large number of message-passing iterations (for example see Figure \ref{fig:Trees}). This can be attributed to the bi-Lipschitzness of FSW-GNN, in contrast to standard MPNNs, which are not bi-Lipschitz \cite{davidson2024h}.

\subsection{Related Works}
\label{subsec_related_works}
%

%


\paragraph{Bi-Lipschitzness}\label{related_work:bi_lip}
Bi-Lipschitzness arises naturally in many domains, including frames \citep{balan1997stability}, phase retrieval \citep{bandeira2014saving, cheng2021stable}, group-invariant learning \citep{cahill2020complete, cahill2024bilipschitz} and  multisets \citep{amir2023neural, amir2024injective}. 
In the context of MPNNs, upper Lipschitzness was discussed in 
\citep{TMD, levie2023a}.
A recent survey by \citet{morris2024future} identifies \emph{bi-Lipschitzness} as a significant future goal for theoretical GNN research. 

\paragraph{SortMPNN}
First steps towards a Bi-Lipschitz MPNN have recently  been made by \citet{davidson2024h}. Their work analyzes a weaker notion of Lipschitz and H{\"o}lder guarantees---\emph{in expectation} over the model parameters. They show that essentially all popular MPNN models fail to be lower-Lipschitz, but are lower-H{\"o}lder in expectation, with an exponent that grows worse as the MPNN depth increases. In contrast, they propose a novel MPNN, called SortMPNN, and prove that it satisfies this weaker notion of bi-Lipschitzness in expectation. 

As we discuss below, the FSW-GNN satisfies the standard, stronger, notion of bi-Lipschitzness. Interestingly, the 
proof technique we present here for FSW-GNN also applies to SortMPNN, allowing us to establish that SortMPNN is bi-Lipschitz as well. Nonetheless, SortMPNN has a key limitation in its message aggregation mechanism: it handles neighborhoods of different sizes by padding them to a predetermined maximal size. This requires \emph{a priori} knowledge of the maximal neighborhood size in all future input graphs---an inherent constraint that significantly limits its applicability. 
In contrast, FSW-GNN does not share this constraint, since it treats vertex neighborhoods as distributions.
Moreover, due to this padding-based approach, SortMPNN only accommodates neighborhoods that are multisets, thus making it unsuitable for graphs with non-integer edge weights. In contrast, FSW-GNN naturally supports edge weights.

\paragraph{MPNNs with advanced pooling mechanisms}
In addition to SortMPNN, our approach is conceptually related to other MPNNs that replace the basic max- or sum-pooling, used for message aggregation, with more advanced pooling mechanisms, such as sorting \citep{balan2022permutation, DGCNN}, standard deviation \citep{pna}, or Sliced-Wasserstein embeddings via template distributions \citep{kolouriwasserstein}. However, these methods lack the bi-Lipschitzness guarantees that our model provides.

\paragraph{WL-equivalent metrics}
Bi-Lipschitz analysis of MPNNs requires a WL-equivalent graph metric. Several such metrics have been proposed, with notable examples being the \emph{Doubly-Stochastic (DS) metric} (also known as the \emph{tree metric}) \citep{grohe}; the \emph{Tree Mover's Distance (TMD)} \citep{TMD}; and the \emph{WL metric} \citep{wlmetric}.
In this paper, we prove that the graph embeddings computed by our FSW-GNN model are bi-Lipschitz with respect to both the DS and TMD metrics.
This analysis applies to graphs of bounded size with continuous, bounded node-features.
Weaker notions of equivalence between these metrics, in the context of graphs with unbounded cardinality and without node features, are discussed in \citep{boker2024fine,boker2021graph}.

\begin{table*}
    \caption{Learning accuracy comparison across different benchmarks and models.}
    \label{tab:performance_comparison}
    \centering
    \begin{tabular}{lccccccccc}
        \toprule
        Model      & Cora  & Cite. & Pubm. & Cham. & Squi. & Actor & Corn. & Texa. & Wisc. \\
        \midrule
        GCN  & 85.77 & 73.68 & \underline{88.13} & 28.18 & 23.96 & 26.86 & 52.70 & 52.16 & 45.88 \\
        GAT & \textbf{86.37} & \underline{74.32} & 87.62 & 42.93 & 30.03 & \underline{28.45} & \underline{54.32} & \underline{58.38} & \underline{49.41} \\
        FSW-GNN & \underline{86.35}  & \textbf{75.44} & \textbf{88.17} & \underline{51.18} & \underline{  36.38} & \textbf{34.66} & \textbf{72.43}  & \textbf{75.68} & \textbf{81.56} \\
        Sort-MPNN & 83.46 & 72.69 & 85.15 & \textbf{78.11} & \textbf{74.69} & 31.32 & 67.03 & 70.54 & 73.92 \\
        \hline
        SOTA & \textbf{90.16} & \textbf{82.07} & \textbf{91.31} & \textbf{79.71} & \textbf{76.71} & \textbf{51.81} & \textbf{92.72} & \textbf{88.38} & \textbf{94.99} \\
        \bottomrule
    \end{tabular}
\end{table*}

\section{Problem Setting}
\label{sec_problem_setting}
In this section, we outline the problem setting, first providing the theoretical background of the problem and then stating our objectives.

\paragraph{Vertex-featured graphs}
Our main objects of study are graphs with vertex features, represented as triplets $G=\br{V,E,X}$, where $V=\brc{v_i}_{i=1}^n$ is the set of vertices, $E\subseteq \setst{ \brc{v_i,v_j}}{i,j\in\brs{n}}$ are the undirected edges in $G$, and $X=\brs{\bx_1,\ldots,\bx_n}$ is a matrix containing the vertex feature vectors $\bx_i\in\Omega$, with  $\Omega \subseteq \RR^d$ being the \emph{feature domain}. We denote by $\GNof\Omega$ the set of all vertex-featured graphs with at most $N$ vertices and corresponding features in $\Omega$. Throughout the paper, we use $\brc{}$ to denote multisets. We note that our results readily extend to graphs with edge features.\ta{; see XX. (to do)}
%

%

\paragraph{Weisfeiler-Lemann Graph Isomorphism test}
Two graphs are \emph{isomorphic} if they are identical up to relabeling of their nodes. Perhaps surprisingly, the problem of determining whether two given graphs are isomorphic is rather challenging. To date, no known algorithm can solve it in polynomial time \citep{babai2016graph}. However, there exist various heuristics that provide an incomplete but often adequate method to test whether a given pair of graphs is isomorphic. The most notable example is the Weisfeiler-Leman (WL) graph isomorphism test.

The WL test can be described as assigning to each graph $G=\br{V,E,X}$ a feature vector $c_G^T$ according to the formula
\begin{equation}
\begin{split}\label{eqdef_wl}
    c^{0}_{v} \eqdefx & \X_{v}, \quad v \in V; 
    \quad \textup{for}\ \ 1 \leq t \leq T:
    \\ c^{t}_{v} \eqdefx & \textup{Combine}\br{c^{t-1}_{v},\textup{Aggregate}\br{\setst{c^{t-1}_{u}}{u \in \N_{v}}}};
    \\ c^{T}_G \eqdefx & \textup{Readout}\br{\brc{c^{T}_{v_{1}},\ldots,c^{T}_{v_{n}}}},
\end{split}
\end{equation}
where Aggregate and Readout are functions that injectively map multisets of vectors in Euclidean space into another Euclidean space, Combine is an injective function from one Euclidean space to another, and $\N_{v}$ denotes the neighborhood of the vertex $v$ in $G$. 

\begin{definition*}[WL graph equivalence]
Two vertex-featured graphs $G$ and $\tilde{G}$ are said to be \emph{WL-equivalent}, denoted by $G \wleq \tilde G$, if $c^T_G = c^T_{\tilde G}$ for all $T \geq 0$. Otherwise, they are said to be \emph{WL-separable}.
\end{definition*}

It is a known fact \citep{grohe2021logic,morris2023weisfeiler} that for $G,\tilde G \in \GNd$, if the equality $c^T_G = c^T_{\tilde G}$ is satisfied for $T=N$, then it is satisfied for all $T \geq 0$, and thus $G \wleq \tilde G$.

While the WL test can separate most pairs of non-isomorphic graphs, there exist examples of non-isomorphic graph pairs that WL cannot separate; see \citep{fail_example}.
%

\paragraph{Message passing neural networks}
Message Passing Neural Networks (MPNNs) 
operate on a similar principle to the WL test, but with the purpose of performing predictions on single graphs rather than determining if pairs of them are isomorphic. 
Their core mechanism is the message-passing procedure, which maintains a hidden feature for each vertex and iteratively updates it as a function of the neighbors' features.
This process is outlined as follows:

\begin{enumerate}
    \item \textbf{Initialization: } The hidden feature $\bh^{(0)}_v$ of each node is initialized by its input feature $\bx_v$.
    \item \textbf{Message aggregation:} Each node $v\in V$ aggregates messages from its neighbors by
    \begin{equation}
        \bmsg_v^{\br{t}} \eqdef \textup{Aggregate}\of{ \setst{ \bh^{\br{t-1}}_u } {u \in \N_v} } 
    \end{equation}
  Where Aggregate is a multiset-to-vector function.   %
    \item \textbf{Update step:} Each node updates its own hidden feature according to its aggregated messages and its previous hidden feature, using a vector-to-vector \emph{update function}:
    \begin{equation}
        \bh^{\br{t}}_v \eqdef \textup{Update}\of{ \bmsg_v^{\br{t}}, \bh^{\br{t-1}}_v },
    \end{equation}
    \item \textbf{Readout:} After $T$ iterations of steps 2-3, a graph-level feature $\bh_G$ is computed by a multiset-to-vector \emph{readout function}:
    \begin{equation*}
        \bh_G \eqdef \textup{Readout}\of{\setst{ \bh^{\br{T}}_v}{v\in V}}.
    \end{equation*}
\end{enumerate}
Numerous MPNNs were proposed in recent years, including GIN \citep{xu2018how}, GraphSage \citep{hamilton17graphsage}, GAT \citep{Velickovic18gat}, and GCN \citep{kipf2017semisupervised},  the main differences between them being the specific choices of the aggregation, update, and readout functions. 

An MPNN computes an \emph{embedding} of a graph $G$ to a vector $F\of{G}= \bh_G$. The obtained embedding is often further processed by standard machine-learning tools for vectors, such as multi-layer perceptrons (MLPs), to obtain a final graph prediction.  
The ability of such a model to approximate functions on graphs is closely related to the separation properties of $F$: if $F$ can differentiate between any pair of non-isomorphic graphs, then a model of the form $\mathrm{MLP}\circ F$ would be able to approximate any functions on graphs \cite{chen2019equivalence}.

Unfortunately, MPNN cannot separate any pair of WL-equivalent graphs, even if they are not truly isomorphic \citep{xu2018how,morris2019WL}. Accordingly, the best we can hope for from an MPNN, in terms of separation, is \emph{WL equivalence}: for every pair of graphs $G,G'\in \GNof\Omega$, $F(G)=F(G')$ \emph{ if and only if } $G \wleq G'$. While MPNNs based on max- or mean-pooling cannot be WL-equivalent \citep{xu2018how}, it is possible to construct WL-equivalent MPNNs based on sum-pooling, as discussed in \citep{xu2018how,morris2019WL,aamand2022exponentially,amir2023neural,bravo2024dimensionality}. 
Theoretically, a properly tuned graph model based on a WL-equivalent MPNN should be capable of perfectly solving any binary classification task, provided that no two WL-equivalent graphs have different ground-truth labels. However, this separation does not always manifest in practice. One reason is that WL-equivalent functions may map two input graphs that are far apart in the input space to outputs that are numerically indistinguishable in the output Euclidean space. In fact, \citet{davidson2024h} provide an example of graph pairs that are not WL-equivalent, yet are mapped to near-identical outputs by standard sum-based MPNNs. Consequently, these MPNNs fail on binary classification tasks for such graphs.

This paper aims to address this limitation by devising an MPNN whose embeddings preserve distances in the bi-Lipschitz sense. To state our goal formally, we first need to define a notion of distance on the input space of graphs.

\paragraph{WL metric for graphs}
WL metrics quantify the \emph{extent} to which two graphs are not WL-equivalent:
\begin{definition}[WL metric]
    \label{WL_metric}
    A \emph{WL metric on $\GNof\Omega$} is a function $\rho:\GNof\Omega \times \GNof\Omega \to \RR_{\geq 0}$ that satisfies the standard requirements for a metric, with the exception that 
    $\rho\of{G_1, G_2} =0 $ if and only if $G_1 \wleq G_2$.
\end{definition}

For convenience, we use the term WL metric, despite the fact that strictly speaking, WL metrics are \emph{pseudometrics} on $\GNd$.

%


\paragraph{Tree Mover's Distance}
The first WL metric we consider is the Tree Mover's Distance (TMD), defined in \citep{TMD}. This metric is based on building computation trees $\tree_v^{(t)}$, which simulate the WL procedure used to create the node features $h_v^{(t)}$, and calculating distances 
\begin{equation}\label{eq:TDshort}
\TD(\tree_v^{(t)},\tree_u^{(t)}) 
\end{equation}
recursively between sub-trees using optimal transport.
These node-level distances are zero if, and only if, the features $c_v^{(t)}$ and $c_u^{(t)}$, constructed by the WL test, are equal.  A graph-level distance $\tmd(G,G')$ is obtained by aggregating all node-level distances.  For the full definition, see \cref{app:TMD} and \cite{TMD}. Under the assumption that the feature domain $\Omega$ does not contain the zero vector,   \citet{TMD} proved that $\tmd(G,G') $ is a WL-metric. 

The second  WL metric we consider is the DS metric \citep{grohe2021logic}. Originally, this metric was defined only for featureless graphs of fixed cardinality. In the next section, we extend this metric to the more general case of $\GNd$.

\paragraph{Bi-Lipschitzness}
Once a WL-metric is defined to measure distances between graphs, one can bound the distortion incurred by a graph embedding with respect to that metric, using the notion of bi-Lipschitzness:
\begin{definition*}[Bi-Lipschitz embedding]
    Let $\rho$ be a WL-metric on $\GNof\Omega$. An embedding $\emb:\GNof\Omega\to\RR^m$ is said to be \emph{bi-Lipschitz with respect to $\rho$ on $\GNof\Omega$} if there exist constants $0<c\leq C < \infty$ such that $\forall G_1, G_2 \in \GNof\Omega$,
    \begin{equation}\label{eqdef_bilip}
    \begin{split}
    \scalebox{0.95}{$
        c \cdot \rho\of{G_1,G_2}
        \leq \norm{\emb\of{G_1}-\emb\of{G_2}}_2
        \leq C \cdot \rho\of{G_1,G_2}.$}
    \end{split}
    \end{equation}
    If $\emb$ only satisfies the left- or right-hand side of \cref{eqdef_bilip}, it is said to be \emph{lower-} or \emph{upper-Lipschitz} respectively.
\end{definition*}
%

Bi-Lipschitzness ensures that the embedding maps the input space $\GNof\Omega$ into the output Euclidean space with bounded distortion, with the ratio $\tfrac{C}{c}$ serving as an upper bound on the distortion, akin to the condition number of a matrix. 

\section{Main Contributions}
\label{sec_main_contributions}
In this section, we discuss our main contributions. We begin by defining our generalized DS metric for vertex-featured graphs. We then discuss our proposed MPNN and show that it is bi-Lipschitz with respect to the DS and TMD metrics.

\paragraph{The DS metric}
The DS metric originates from a relaxation of the \emph{graph isomorphism problem}. Two graphs $G$ and $\tilde G$, each with $n$ vertices, and corresponding adjacency matrices  $A$, $\tilde A$ are isomorphic if and only if there exists a \emph{permutation matrix} $P$ such that $AP=P\tilde A $. Since checking whether graphs are isomorphic is intractable, an approximate solution can be sought by considering the equation $AS=S\tilde A$ with $S \in \D_n$, the collection of $n \times n$ \emph{doubly stochastic} matrices, which is the convex hull of all permutation matrices.
Remarkably, this equation admits a solution $S\in\D_n$ if and only if the graphs $G$ and $\tilde G$ are WL-equivalent \citep{book}. Accordingly, a WL-metric between featureless graphs with the same number of vertices $n$ can be defined by the minimization problem
\begin{equation}
    \label{eqdef_metric_ds_nofeat}
    \rhods\of{G,\tilde G}
    =
    \min_{S\in\D_n} \ 
    \norm{AS-S\tilde A},
\end{equation}
where $\norm{\argdot}$ could denote any $p$ norm. The fact that this is indeed a pseudometric was established in \cite{bento}. The optimization problem \cref{eqdef_metric_ds_nofeat} can be solved by off-the-shelf convex optimization solvers and was considered as a method for finding the correspondence between two graphs in several papers, including \cite{aflalo,risk,dym2018exact,dym2017ds++,Bernard_2018_CVPR}. 

The idea of using the DS metric for MPNN stability analysis was introduced in \citep{grohe} and further discussed by \citet{boker2021graph}. To apply this idea to our setting, we need to adapt this metric to vertex-featured graphs with varying numbers of vertices. We do this by augmenting it as follows:
\begin{equation}\label{eqdef_metric_ds}
    \rhods\of{G,\tilde G}
    = 
    \abs{n-\tilde n}
    +
    \min_{S\in\Pi\of{n,\tilde n}} \ 
    \norm{AS-S\tilde A}_1
     + 
    \sum_{i\in\brs{n}, j\in\brs{\tilde n}} S_{ij}
    \norm{\bx_i-\btx_j}_1,
\end{equation}
where $n$ and $\tilde{n}$ denote the number of vertices in $G$ and $\tilde{G}$, $\bx_i$ and $\btx_j$ denote the vertex features of $G$ and $\tilde G$, and   $\Pi\of{n,\tilde n}$ is the set of $n \times \tilde n$ matrices $S$ with non-negative entries, whose rows and columns sum to $n$ and $\tilde n$, respectively. 
%

%
\begin{restatable}{theorem}{thmRhoDSIsWLMetric}
	\label{thm_rhods_is_wl_metric}
	\onmain{\textup{(Proof in \cref{subapp_proofs_ds_metric})\ }}
$\rhods$ is a WL-equivalent metric on $\GNd$.
\end{restatable}



%

%
\paragraph{Bi-Lipschitz MPNN} 
We now present our main contribution: a novel MPNN that is not only WL-equivalent, but also bi-Lipschitz, with respect to both $\rhods$ and TMD.

\paragraph{FSW Embedding} The core innovation in our MPNN lies in its message aggregation method. Specifically, we employ the \emph{Fourier Sliced-Wasserstein (FSW) Embedding}---a method for embedding multisets of vectors into Euclidean space, proposed by \citet{amir2024injective}, where it was also shown to be bi-Lipschitz. This property makes it plausible, a priori, that an MPNN based on FSW aggregation will be bi-Lipschitz for \emph{graphs}. In this work, we formally establish that this is indeed the case. We begin by describing the FSW embedding and then introduce our FSW-GNN architecture.

The FSW embedding maps an input multiset $ \bX = \brc{\bx_1,\ldots,\bx_n}$, where $\bx_1,\ldots,\bx_n \in \RR^d$, to an output vector $\bz = \br{z_1,\ldots,z_m} \in \RR^m$. 
It is denoted by
\begin{equation*}
\bz 
=
\embfsw\of{\bX; \br{\bv_k,\xi_k}_{k=1}^m},
\quad 
\bX = \brc{\bx_1,\ldots,\bx_n}.
\end{equation*}
In addition to the input multiset $\bX$, it depends on parameters $\br{\bv_k, \xi_k}_{k=1}^{m-1}$, where each $\bv_k \in \S^{d-1}$ represents a direction vector and $\xi_k \in \RR$ represents a frequency. The embedding is computed in three steps: First, the direction $\bv_k$ is used to project the original multiset of vectors to  a multiset of scalars 
$\brc{\bra{\bv_k,\bx_i}}_{i=1}^n$,
which is then sorted: $\by_k = \sort\of{\brc{\bra{\bv_k,\bx_i}}_{i=1}^n}$. This step is similar to the sort-type embedding used in SortMPNN, and was shown to be bi-Lipschitz on multisets of fixed size \citep{balan2022permutation}. However, taking $\by_k$ directly as the embedding leads to an output dimension dependent on the input multiset's size, thus making the embedding unsuitable for varying-size multisets. The next steps address and resolve this limitation.

In the second step, the vector $\by_k$ is identified with a step function $Q_{\by_k}: \brs{0,1}\to\RR$, namely the \emph{quantile function} of the multiset $\brc{\bra{\bv_k,\bx_i}}_{i=1}^n$; see illustration in \cref{fig_quantile_function}. Then, in the third step, the \emph{cosine transform}, a variant of the Fourier transform, is applied to $Q_{\by_k}$, and sampled at the given frequency $\xi_k$, to obtain the final output coordinates $z_k$, $k=1,\ldots,m-1$. This is summarized by:
%
\begin{wrapfigure}[12]{r}{0.3\linewidth}
    \centering
    \includegraphics[width=\linewidth]{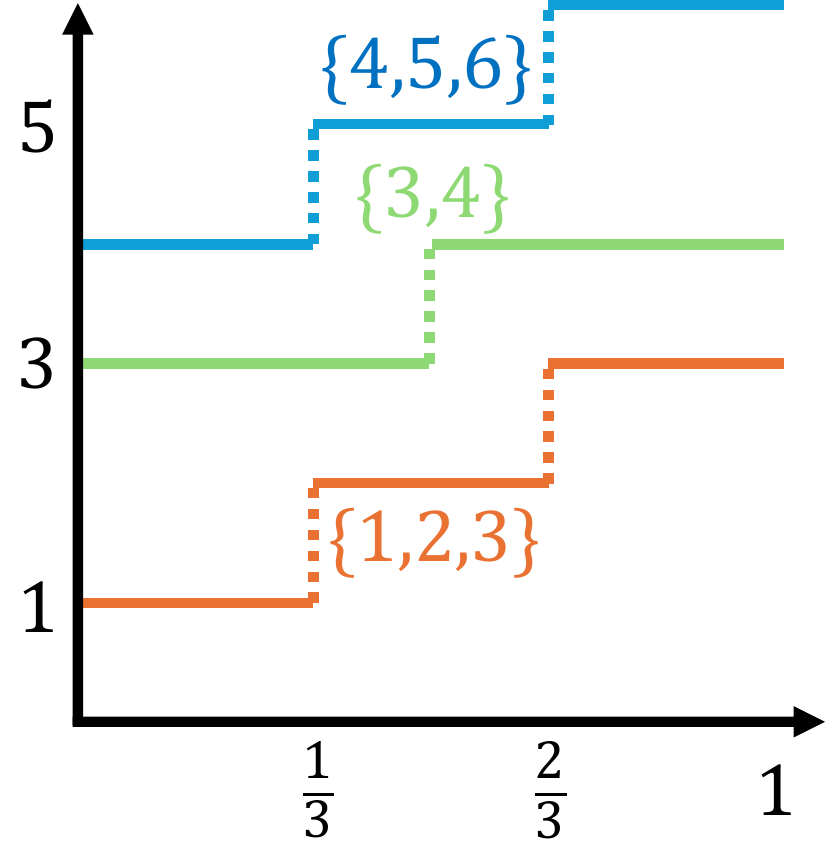}     \caption{\label{fig_quantile_function}The quantile function for three different multisets}      
\end{wrapfigure}
\begin{align}
\by_k
&=
\resizebox{!}{0.9\height}{$\br{y_{k1},\ldots,y_{kn}}$}
=
\mathrm{sort}\br{\resizebox{!}{0.85\height}{$\bra{\bv_k, \bx_1},\ldots, \bra{\bv_k, \bx_n}$} } \label{eq:sort}\\
Q_{\by_k}(t)
&=
%
%
y_{ki} \ \textup{with $i$ such that}\ t \in [\tfrac{i-1}{n},\tfrac{i}{n}) \label{eq:quantile}
\\
z_k&=2(1+\xi_k)\int_0^1 Q_{\by_k}(t) \cos(2\pi \xi_k t)dt \label{eq:cosine}
\end{align}

Lastly, note that \cref{eq:sort,eq:cosine,eq:quantile} treat input multisets as uniform distributions over their elements, and thus are agnostic to the multiset size by design. To address this and ensure that multisets of different sizes but identical element-proportions are mapped to distinct outputs, the last output coordinate $z_m$ is set to the cardinality $\abs{\bX}$ of the input multiset $\bX$. Further details appear in \citep[Appendix A.1]{amir2024injective}.

Note that the integral in \cref{eq:cosine} has a closed-form solution, and the whole embedding can be computed with a complexity of $\mathcal{O}\of{m n d + m n \log n}$, similarly to simple aggregation functions such as sum pooling. Furthermore, the embedding parameters and output dimension do not depend on $n$, making this method suitable for multisets of different sizes.
\paragraph{FSW-GNN} The FSW-GNN model processes input graphs $G=\br{V,E,X}$ by $T$ message-passing iterations: 
\begin{equation}\label{eqdef_fswgnn}
\begin{split}
    \bh^{\br{0}}_{v} \eqdefx & \bx_{v},
    %
    \\ \bq_v^{\br{t}} 
    \eqdefx &
    \embfswt[t]\of{ \setst{ \bh^{\br{t-1}}_u } {u \in \N_v} }, 
    \ \ \ 1 \leq t \leq T,
    \\ \bh^{\br{t}}_v
    \eqdefx &
    \Phi^{\br{t}} \of{ \brs{\bh^{\br{t-1}}_v;\bq^{\br{t}}_v}},
\end{split}
\end{equation}
where the functions $\embfswt[t]$ are all instances of the FSW embedding,
%
$\Phi^{\br{t}}$ are MLPs,
and $\brs{\bx; \by}$ denotes column-wise concatenation of column vectors $\bx$ and $\by$. Finally, a graph-level output is computed by:
\begin{equation}\label{eqdef_fswgnn_readout}
    \bh_G \eqdef 
    \Psi \circ
    \embfswglob\of{\setst{ \bh^{\br{T}}_v}{v\in V}},
\end{equation}
where, again, $\embfswglob$ is an FSW embedding, and $\Psi$ an MLP.

The following \lcnamecref{thm_fswgnn_is_wl_equivalent} shows that, with the appropriate choice of MLP sizes and number of iterations $T$, our proposed architecture is WL equivalent:

\begin{restatable}{theorem}{thmFSWGNNIsWLEquivalent}\label{thm_fswgnn_is_wl_equivalent}\onmain{\textup{(Proof in \cref{subapp_proofs_fswgnn_wl_equivalence})}}
    Consider the FSW-GNN architecture for input graphs in $\GNd$, with $T=N$ iterations, where $\Phi^{\br{t}}, \Psi$  are just linear funtions, and all features (except for input features) are of dimension $m \geq 2Nd+2$. Then for Lebesgue almost every choice of model parameters, the graph embedding defined by the architecture is WL equivalent. 
\end{restatable}


%

The proof of \cref{thm_fswgnn_is_wl_equivalent} is based on the theory of $\sigma$-subanalytic functions and the \emph{Finite Witness Theorem}, introduced in \citep{amir2023neural}.

It is worth noting that the output dimension $m$ required in practice is typically considerably lower than the one required in \cref{thm_fswgnn_is_wl_equivalent}. This can be explained intuitively by the following fact: if all input graphs originate from a subset of $\GNd$ with intrinsic dimension $D$ that is lower than the ambient dimension $n\cdot d$, then it can be shown that $m=2D+2$ suffices for WL-equivalence.

\paragraph{From separation to bi-Lipschitzness}
In general, WL-equivalence does not imply bi-Lipschitzness. As mentioned above, sum-based MPNN can be injective but are never bi-Lipschitz. In contrast, we shall now prove that for FSW-GNN,
 WL-equivalence does imply bi-Lipschitzness, under the assumption that the feature domain $\Omega$ is compact:

\begin{restatable}{theorem}{thmFSWGNNIsBilip}\label{thm_fswgnn_is_bilip}
\onmain{\textup{(Proof in \cref{subapp_proofs_fswgnn_bilipschitz})}}
Let $\Omega \subset \RR^d$ be compact. 
Under the assumptions of
\cref{thm_fswgnn_is_wl_equivalent}, the FSW-GNN is bi-Lipschitz with respect to $\rhods$ on $\GNof\Omega$. If, additionally, $\Omega$ does not contain $\bzero$, then the FSW-GNN is bi-Lipschitz with respect to TMD on $\GNof\Omega$. 
\end{restatable}

We now give a high-level explanation of the proof idea. We rely on the following facts: (1) the output of FSW-GNN for an input graph $G=\br{V,E,X}$ is piecewise-linear with respect to the vertex-feature matrix $X$. This follows from properties of the FSW embedding functions $\embfswt[t]$ and $\embfswglob$ used in \cref{eqdef_fswgnn,eqdef_fswgnn_readout}. (2) both metrics $\rhods$ and TMD can be transformed, with bounded distortion, into metrics that are piecewise-linear, by choosing all the vector norms they employ to be the $\ell_1$ norm. The claim then follows from these observations and the following \lcnamecref{thm_pwl_ratio_bound}, which shows that two functions that are piecewise linear and have the same zero set, are bi-Lipschitz with respect to one another:
\begin{restatable}{lemma}{thmPWLRatioBound}
\label{thm_pwl_ratio_bound}
    \onmain{\textup{(Proof in \cref{subapp_proofs_fswgnn_bilipschitz})}}
    Let $f,g: M \to \RR_{\geq 0}$ be nonnegative piecewise-linear functions defined on a compact polygon $M \subset \RR^d$. Suppose that for all $x\in M$, $f(x) = 0$ if and only if $g(x)=0$. Then there exist real constants $c,C > 0$ such that
    \begin{equation}\label{eq_pwl_functions_ratio_bound}
        c \cdot g\of{x} \leq f\of{x} \leq C \cdot g\of{x}, \quad \forall x \in M.
    \end{equation}
\end{restatable}
We note that the assumption of a compact domain is essential; see Remark~6 in \cite{balan2024stability}. While \cref{thm_pwl_ratio_bound} is rather intuitive, we are not aware of it appearing previously in the literature. This lemma easily implies\footnotemark\ previous bi-Lipschitzness results in the literature, such as Theorem~1.4 in \citep{balan2023ginvariant}, Theorem~1 in \citep{balan2024stability}, and Theorem~3.10 in \citep{balan2022permutation}. We believe the lemma has the potential to serve as a valuable tool for proving bi-Lipschitzness results in additional domains in the future. To illustrate this, in Appendix \ref{app:subgraph} we show how the lemma can be used to construct a bi-Lipschitz \emph{subgraph aggregation MPNN}.

\footnotetext{In all these scenarios, the proposed embedding and metric are \emph{homogeneous}, so that bi-Lipschitzness on a compact polygon implies global bi-Lipschitzness.}

To complement \cref{thm_fswgnn_is_bilip}, we now present an analogous result for node-level tasks: we show that the node features computed by FSW-GNN are bi-Lipschitz with respect to the \emph{Tree Distance} discussed in \cref{eq:TDshort}. 
\begin{restatable}{theorem}{thmNodeLevelBiLip}\label{thm_node_level_bilip}
\onmain{\textup{(Proof in \cref{subapp_proofs_fswgnn_bilipschitz})}}
Suppose that $\Omega \subset \RR^d$ is a compact set that does not contain zero.
Under the assumptions of
\cref{thm_fswgnn_is_wl_equivalent}, the node features computed by FSW-GNN are bi-Lipschitz with respect to the Tree Distance metric on $\GNof\Omega$. 
\end{restatable}

\section{Oversmoothing, Oversquashing, and Bi-Lipschitz MPNN}\label{sec:over}
Training \emph{deep} MPNNs is one of the core challenges in graph neural networks \citep{morris2024future}. The difficulty is often attributed to \emph{oversmoothing} \citep{Rusch2023ASO} or \emph{oversquashing} \citep{alon}. Both phenomena can be regarded as a form of metric distortion induced by the MPNN as the number of iterations  grows. Oversmoothing is the phenomenon where, for large $t$, 
\begin{equation}\label{eq:oversmoothing}
\norm{\bx_v^{(t)}- \bx_u^{(t)}} \approx 0, \quad \forall v,u\in V.
\end{equation}
Oversquashing, as defined in \cite{bron}, refers the phenomenon where a non-negligible change in an input node feature $\bx_v^{(0)}$ to a new value $\hat{\bx}_v^{(0)}$ results in only a negligible change 
in features of far-away nodes, namely, 
\begin{equation}\label{eq:oversquashing}
	\norm{ \bx_u^{(t)}- \hat{\bx}_u^{(t)} }  \ll \norm{ \bx_v^{(0)}- \hat{\bx}_v^{(0)} }, \text{ for } u \text{\ far from\ } v.
\end{equation}
In other words, information fails to propagate adequately between distant nodes through the message-passing process.

Accordingly, an MPNN that perfectly preserves distances between node features would provably not encounter oversmoothing or oversquashing. We note that FSW-GNN is not in this position: while it is bi-Lipschitz and thus has bounded distortion, we cannot rule out the possibility that its distortion grows with $t$. Nonetheless, we conjecture that FSW-GNN should perform much better than standard MPNN on long range tasks. This is because, as shown in \cite{davidson2024h}, standard MPNNs are only lower H\"{o}lder in expectation, namely, the inequality
$$\|x_v^{(t)}- x_u^{(t)}\| \geq C_t \left[\TD(\tree_v^{(t)},\tree_u^{(t)}) \right]^{\alpha_t} $$ 
holds only in expectation, 
with constants $C_t$ and exponents $\alpha_t$ that grow with $t$. In contrast, for FSW-GNN we have $\alpha_t=1$ for all $t$, and only $C_t$ may grow with $t$. 
We provide empirical evidence for our conjecture in \cref{sec_numerical_experiments}, where we show that our bi-Lipschitz MPNN outperforms standard MPNNs on long-range tasks.
%
%

\section{Numerical Experiments}
\label{sec_numerical_experiments}
We compare the performance of FSW-GNN with standard MPNNs and Sort-MPNN on both real-world benchmarks and synthetic long-range tasks\footnote{Our code will be available to the public upon paper acceptance.}. 

\begin{figure*}
    \centering
    \begin{minipage}[b]{0.32\textwidth}
        \centering
        \definecolor{darkgreen}{rgb}{0.0, 0.5, 0.0} 
\definecolor{brown}{rgb}{0.6, 0.4, 0.2} 
\definecolor{teal}{rgb}{0.0, 0.5, 0.5} 

\begin{tikzpicture}
\begin{axis}[
    legend to name=LegendOversquashing, 
    width=\linewidth, 
    height=0.85\linewidth, 
    xlabel={Number of MPNN Layers},
    xlabel style={font=\scriptsize},
    ylabel style={font=\scriptsize},
    xmin=1, xmax=16,
    ymin=10, ymax=110,
    legend style={
        at={(0.04,0.04)}, 
        anchor=south west, 
        font=\small,
        nodes={scale=1, transform shape}, 
        legend columns=6 
    },
    grid=major,
    grid style={opacity=0.5},
    tick label style={
        /pgf/number format/fixed, 
        /pgf/number format/precision=2, 
        /pgf/number format/zerofill=false, 
        font=\scriptsize,
    },
    xtick={2,4,6,8,10,12,14},
    ytick={0, 20, 40, 60, 80, 100},
    minor x tick num=1,
    tick style={
        major tick length=5pt, 
        minor tick length=2pt, 
    },
    extra x ticks={15},        
    extra x tick style={
        tick style={thin},
        major tick length=2pt, 
        grid=none, 
        tick label style={opacity=0} 
    },  
    yticklabel={\pgfmathprintnumber{\tick}\kern0.1em\%},
    scaled ticks=false, 
    enlargelimits=false, 
    axis line style={}, 
    clip=false 
]

\addplot[
    color=orange, 
    mark=square*, 
    mark size=2pt,
    mark options={scale=0.7},
    line width=1pt,
    solid]
    table[x=Number of MPNN Layers, y=FSW-GNN (Ours), col sep=comma] {results_LongRange_CliquePath.csv};
\addlegendentry{FSW-GNN}

\addplot[
    color=red, 
    mark=triangle*, 
    mark size=3pt,
    mark options={scale=0.7},
    line width=1pt,
    dashed]
    table[x=Number of MPNN Layers, y=GIN, col sep=comma] {results_LongRange_CliquePath.csv};
\addlegendentry{GIN}

\addplot[
    color=purple, 
    mark=diamond*, 
    mark size=3pt,
    mark options={scale=0.7},
    line width=1pt,
    dotted]
    table[x=Number of MPNN Layers, y=GAT, col sep=comma] {results_LongRange_CliquePath.csv};
\addlegendentry{GAT}

\addplot[
    color=blue, 
    mark=*, 
    mark size=2pt,
    mark options={scale=0.7},
    line width=1pt,
    solid]
    table[x=Number of MPNN Layers, y=GCN, col sep=comma] {results_LongRange_CliquePath.csv};
\addlegendentry{GCN}

\addplot[
    color=brown, 
    mark=square*, 
    mark size=2pt,
    mark options={scale=0.7},
    line width=1pt,
    dash dot]
    table[x=Number of MPNN Layers, y=Sort-MPNN, col sep=comma] {results_LongRange_CliquePath.csv};
\addlegendentry{Sort-MPNN}

\addplot[
    color=teal, 
    mark=star, 
    mark size=3pt,
    mark options={scale=0.7},
    line width=1pt,
    dash dot]
    table[x=Number of MPNN Layers, y=SAGE, col sep=comma] {results_LongRange_CliquePath.csv};
\addlegendentry{Graph-SAGE}

\end{axis}
\end{tikzpicture}
        \vspace{-10pt}
        \caption*{\small (a) CliquePath}
        \label{fig:cliquepath}
    \end{minipage}
    \hfill
    \begin{minipage}[b]{0.32\textwidth}
        \centering
        \definecolor{darkgreen}{rgb}{0.0, 0.5, 0.0} 
\definecolor{brown}{rgb}{0.6, 0.4, 0.2} 
\definecolor{teal}{rgb}{0.0, 0.5, 0.5} 

\begin{tikzpicture}
\begin{axis}[
    legend to name=LegendRing, 
    width=\linewidth, 
    height=0.85\linewidth, 
    xlabel={Number of MPNN Layers},
    xlabel style={font=\scriptsize},
    ylabel style={font=\scriptsize},
    xmin=1, xmax=16,
    ymin=10, ymax=110,
    legend style={
        at={(0.04,0.04)}, 
        anchor=south west, 
        font=\small,
        nodes={scale=0.05, transform shape}, 
        legend columns=1 
    },
    grid=major,
    grid style={opacity=0.5},
    tick label style={
        /pgf/number format/fixed, 
        /pgf/number format/precision=2, 
        /pgf/number format/zerofill=false, 
        font=\scriptsize,
    },
    xtick={2,4,6,8,10,12,14},
    ytick={0, 20, 40, 60, 80, 100},
    minor x tick num=1,
    tick style={
        major tick length=5pt, 
        minor tick length=2pt, 
    },
    extra x ticks={15},        
    extra x tick style={
        tick style={thin},
        major tick length=2pt, 
        grid=none, 
        tick label style={opacity=0} 
    },  
    yticklabel={\pgfmathprintnumber{\tick}\kern0.1em\%},
    scaled ticks=false, 
    enlargelimits=false, 
    axis line style={}, 
    clip=false 
]

\addplot[
    color=orange, 
    mark=square*, 
    mark size=2pt,
    mark options={scale=0.7},
    line width=1pt,
    solid]
    table[x=Number of MPNN Layers, y=FSW-GNN (Ours), col sep=comma] {results_LongRange_Ring.csv};
\addlegendentry{FSW-GNN}

\addplot[
    color=red, 
    mark=triangle*, 
    mark size=3pt,
    mark options={scale=0.7},
    line width=1pt,
    dashed]
    table[x=Number of MPNN Layers, y=GIN, col sep=comma] {results_LongRange_Ring.csv};
\addlegendentry{GIN}

\addplot[
    color=purple, 
    mark=diamond*, 
    mark size=3pt,
    mark options={scale=0.7},
    line width=1pt,
    dotted]
    table[x=Number of MPNN Layers, y=GAT, col sep=comma] {results_LongRange_Ring.csv};
\addlegendentry{GAT}

\addplot[
    color=blue, 
    mark=*, 
    mark size=2pt,
    mark options={scale=0.7},
    line width=1pt,
    solid]
    table[x=Number of MPNN Layers, y=GCN, col sep=comma] {results_LongRange_Ring.csv};
\addlegendentry{GCN}

\addplot[
    color=brown, 
    mark=square*, 
    mark size=2pt,
    mark options={scale=0.7},
    line width=1pt,
    dash dot]
    table[x=Number of MPNN Layers, y=Sort-MPNN, col sep=comma] {results_LongRange_Ring.csv};
\addlegendentry{Sort-MPNN}

\addplot[
    color=teal, 
    mark=star, 
    mark size=3pt,
    mark options={scale=0.7},
    line width=1pt,
    dash dot]
    table[x=Number of MPNN Layers, y=SAGE, col sep=comma] {results_LongRange_Ring.csv};
\addlegendentry{SAGE}

\end{axis}
\end{tikzpicture}
        \vspace{-10pt}
        \caption*{\small (b) Ring}
        \label{fig:ring}
    \end{minipage}
    \hfill
    \begin{minipage}[b]{0.32\textwidth}
        \centering   
        \definecolor{darkgreen}{rgb}{0.0, 0.5, 0.0} 
\definecolor{brown}{rgb}{0.6, 0.4, 0.2} 
\definecolor{teal}{rgb}{0.0, 0.5, 0.5} 

\begin{tikzpicture}
\begin{axis}[
    legend to name=LegendCrossRing, 
    width=\linewidth, 
    height=0.85\linewidth, 
    xlabel={Number of MPNN Layers},
    xlabel style={font=\scriptsize},
    ylabel style={font=\scriptsize},
    xmin=1, xmax=16,
    ymin=10, ymax=110,
    legend style={
        at={(0.04,0.04)}, 
        anchor=south west, 
        font=\small,
        nodes={scale=0.05, transform shape}, 
        legend columns=1 
    },
    grid=major,
    grid style={opacity=0.5},
    tick label style={
        /pgf/number format/fixed, 
        /pgf/number format/precision=2, 
        /pgf/number format/zerofill=false, 
        font=\scriptsize,
    },
    xtick={2,4,6,8,10,12,14},
    ytick={0, 20, 40, 60, 80, 100},
    minor x tick num=1,
    tick style={
        major tick length=5pt, 
        minor tick length=2pt, 
    },
    extra x ticks={15},        
    extra x tick style={
        tick style={thin},
        major tick length=2pt, 
        grid=none, 
        tick label style={opacity=0} 
    },  
    yticklabel={\pgfmathprintnumber{\tick}\kern0.1em\%},
    scaled ticks=false, 
    enlargelimits=false, 
    axis line style={}, 
    clip=false 
]

\addplot[
    color=orange, 
    mark=square*, 
    mark size=2pt,
    mark options={scale=0.7},
    line width=1pt,
    solid]
    table[x=Number of MPNN Layers, y=FSW-GNN (Ours), col sep=comma] {results_LongRange_CrossRing.csv};
\addlegendentry{FSW-GNN}

\addplot[
    color=red, 
    mark=triangle*, 
    mark size=3pt,
    mark options={scale=0.7},
    line width=1pt,
    dashed]
    table[x=Number of MPNN Layers, y=GIN, col sep=comma] {results_LongRange_CrossRing.csv};
\addlegendentry{GIN}

\addplot[
    color=purple, 
    mark=diamond*, 
    mark size=3pt,
    mark options={scale=0.7},
    line width=1pt,
    dotted]
    table[x=Number of MPNN Layers, y=GAT, col sep=comma] {results_LongRange_CrossRing.csv};
\addlegendentry{GAT}

\addplot[
    color=blue, 
    mark=*, 
    mark size=2pt,
    mark options={scale=0.7},
    line width=1pt,
    solid]
    table[x=Number of MPNN Layers, y=GCN, col sep=comma] {results_LongRange_CrossRing.csv};
\addlegendentry{GCN}

\addplot[
    color=brown, 
    mark=square*, 
    mark size=2pt,
    mark options={scale=0.7},
    line width=1pt,
    dash dot]
    table[x=Number of MPNN Layers, y=Sort-MPNN, col sep=comma] {results_LongRange_CrossRing.csv};
\addlegendentry{Sort-MPNN}

\addplot[
    color=teal, 
    mark=star, 
    mark size=3pt,
    mark options={scale=0.7},
    line width=1pt,
    dash dot]
    table[x=Number of MPNN Layers, y=SAGE, col sep=comma] {results_LongRange_CrossRing.csv};
\addlegendentry{SAGE}

\end{axis}
\end{tikzpicture}
        \vspace{-10pt}
        \caption*{\small (c) CrossRing}
        \label{fig:crossring}
    \end{minipage}
    %
    
    \vspace{5pt}
    \pgfplotslegendfromname{LegendOversquashing}
    \caption{\small Performance comparison of MPNN models across the CliquePath, Ring, and CrossRing graph transfer tasks as presented in \cite{bron}.}
    \label{fig:comparison}
\end{figure*}

\paragraph{Empirical distortion evaluation}
\begin{figure*}[t]
    \centering
    \begin{minipage}{0.45\textwidth}
        \centering
        \definecolor{darkgreen}{rgb}{0.0, 0.5, 0.0} 

\begin{tikzpicture}
\begin{axis}[
    width=\linewidth, 
    height=0.85\linewidth, 
    xlabel={Number of MPNN iterations},
    ylabel={Distortion},
    xlabel style={font=\small},
    ylabel style={font=\small},
    legend style={
        at={(0.75,0.125)}, 
        anchor=south, 
        font=\small,
        nodes={scale=0.75, transform shape}, 
        legend columns=1 
    },
    xmin=0.5, xmax=9,
    ymin=0.1, ymax=1000000000000000,
    grid=both,
    grid style={opacity=0.5},
    ymode=log, 
    log basis y=10,
    xtick=data,
    ytick={1, 100, 10000, 1000000, 100000000, 10000000000, 1000000000000, 100000000000000},
    yticklabels={$1$, $10^2$, $10^4$, $10^6$, $10^8$, $10^{10}$, $10^{12}$, $10^{14}$},
    tick label style={/pgf/number format/fixed}
]

\addplot[
    color=blue, 
    mark=*, 
    mark size=2pt, 
    line width=1pt, 
    solid 
] table[x=Layers, y=GCN, col sep=comma] {empirical_distortion_vs_ds.csv};
\addlegendentry{GCN}

\addplot[
    color=red, 
    mark=triangle*, 
    mark size=3pt, 
    line width=1pt, 
    dashed 
] table[x=Layers, y=GIN, col sep=comma] {empirical_distortion_vs_ds.csv};
\addlegendentry{GIN}

\addplot[
    color=purple, 
    mark=diamond*, 
    mark size=3pt, 
    line width=1pt, 
    dotted 
] table[x=Layers, y=GAT, col sep=comma] {empirical_distortion_vs_ds.csv};
\addlegendentry{GAT}

\addplot[
    color=orange, 
    mark=square*, 
    mark size=2pt, 
    line width=1pt, 
    dash dot 
] table[x=Layers, y=Sort-MPNN, col sep=comma] {empirical_distortion_vs_ds.csv};
\addlegendentry{Sort-MPNN}

\addplot[
    color=darkgreen, 
    mark=square*, 
    mark size=2pt, 
    line width=1pt, 
    solid 
] table[x=Layers, y=FSW-GNN, col sep=comma] {empirical_distortion_vs_ds.csv};
\addlegendentry{FSW-GNN}

\end{axis}
\end{tikzpicture}
        \caption{\centering Empirical distortion evaluation with respect to the Doubly-Stochastic (DS) metric}
        \label{fig_empirical_distortion_ds}
    \end{minipage}
    \hfill
    \begin{minipage}{0.45\textwidth}
        \centering
        \vspace{-8pt}
        \definecolor{darkgreen}{rgb}{0.0, 0.5, 0.0} 
\definecolor{brown}{rgb}{0.6, 0.4, 0.2} 
\definecolor{teal}{rgb}{0.0, 0.5, 0.5} 

\begin{tikzpicture}
\begin{axis}[
    width=\linewidth, 
    height=0.85\linewidth, 
    xlabel={Number of MPNN iterations},
    ylabel={MAD energy},
    xlabel style={font=\small},
    ylabel style={font=\small},
    ymin=-0.05, ymax=0.5,
    legend style={
        at={(0.98,0.98)}, 
        anchor=north east, 
        font=\small,
        nodes={scale=0.93, transform shape}, 
        legend columns=2 
    },
    grid=both,
    grid style={opacity=0.5},
    tick label style={
        /pgf/number format/fixed, 
        /pgf/number format/precision=2, 
        /pgf/number format/zerofill=false 
    },
    scaled ticks=false 
]

\addplot[
    color=orange, 
    mark=square*, 
    mark size=2pt,
    line width=1pt,
    solid]
    table[x=Number of MPNN Layers, y=FSW-GNN, col sep=comma] {results_dirichlet.csv};
\addlegendentry{FSW-GNN}

\addplot[
    color=red, 
    mark=triangle*, 
    mark size=3pt,
    line width=1pt,
    dashed]
    table[x=Number of MPNN Layers, y=GIN, col sep=comma] {results_dirichlet.csv};
\addlegendentry{GIN}

\addplot[
    color=teal, 
    mark=star, 
    mark size=3pt,
    line width=1pt,
    dash dot]
    table[x=Number of MPNN Layers, y=Graph-SAGE, col sep=comma] {results_dirichlet.csv};
\addlegendentry{Graph-SAGE}

\addplot[
    color=purple, 
    mark=diamond*, 
    mark size=3pt,
    line width=1pt,
    dotted]
    table[x=Number of MPNN Layers, y=GAT, col sep=comma] {results_dirichlet.csv};
\addlegendentry{GAT}

\addplot[
    color=blue, 
    mark=*, 
    mark size=2pt,
    line width=1pt,
    solid]
    table[x=Number of MPNN Layers, y=GCN, col sep=comma] {results_dirichlet.csv};
\addlegendentry{GCN}

\end{axis}
\end{tikzpicture}        
        \vspace{8pt}
        \caption{MAD Energy vs. Number of MPNN layers for various models on the Ring long-range task}
        \label{fig:oversmoothing}
    \end{minipage}        
\end{figure*}

First, to assess the metric distortion induced by different MPNNs, we compared the distances induced by each MPNN vs. the TMD and DS metric on a particularly challenging set of graph pairs; see \cref{app_numerical_experiments} for details. Empirical estimates $\hat C, \hat c$ of the constants $C$,$c$ of \cref{eqdef_bilip} were computed, and the distortion estimate was taken as the ratio $\hat{C}/\hat{c}$. The results appear in \cref{fig_empirical_distortion_ds}. Similar results were obtained for the TMD; see \cref{fig_empirical_distortion_tmd} in the appendix. As seen in the figures, our method yields considerably lower distortion than all competitors, which aligns with our theoretical guarantees.


\paragraph{Transductive Learning}
Next, we compare FSW-GNN with GCN, GAT and Sort-MPNN for nine transductive learning tasks taken from \cite{geom-gcn}. As shown in  \cref{tab:performance_comparison}, FSW-GNN outperforms the competition in six out of nine tasks. SortMPNN is a clear winner in two of the other graphs, both of which have relatively large average degrees (see Table \ref{tab:common_graph_stats}). In addition, we include the state-of-the-art results known to us for each dataset: For Cora \cite{best_cora}; for Citeseer \cite{best_cite}; for PubMed \cite{best_pubm}; for Chameleon \cite{best_cham}; for Squirrel \cite{best_squirrel}; for Actor \cite{best_actor}; for Cornell \cite{best_corn}; for Texas \cite{best_texas}; for Wisconsin \cite{best_wisc}. These results are typically achieved by models that are strictly more powerful and computationally expensive than any MPNN.

\paragraph{Graph classification and regression}
\vspace{-0.5em} 
In the appendix, we  we include results on standard graph classication and regression tasks: the peptides-func and peptides-struct datasets of the LRGB benchmark \cite{lrgb2022}, and  he MUTAG  and Protein datasets \cite{TUD}. These results appear in \cref{tab:lrgb_datasets}
and  \cref{tab:mutag}.


\paragraph{Long range tasks} Next, we consider several synthetic long-range tasks proposed in the oversquashing literature \cite{alon,bron}. In these problems, one can control the problem radius $r$, which is the number of MPNN iterations needed to solve the problem. 

We first consider the NeighborsMatch problem from
\cite{alon}.  In \cref{fig:Trees}, we compare the performance of  FSW-GNN with standard MPNN on the NeigborsMatch problem with $2 \leq r \leq 8$. Our FSW-GNN achieves perfect accuracy for all values of $r$, while Sort-MPNN fails at $r=8$ and the other competing methods falter for $r\geq 6$. 

Next, we consider the ‘graph transfer’ tasks \citep{bron}, in which a ‘source’ node feature is propagated to a ‘target’ node that is $r$ steps away. All other nodes have ‘blank’ features. Here we used three graph topologies proposed by \cite{bron}: clique, ring, and crossring, with problem radii $r$ varying from $2$ to $15$.

As shown in \cref{fig:comparison}, FSW-GNN is the only method to achieve $100\%$ accuracy across all three graphs and radii. Other models start failing at much smaller $r$, which depends on the graph topology. We do find that our model's performance  deteriorates when $r>20$.

Finally, we evaluate the relation between our empirical results and oversmoothing on the ring experiment, using the Mean Average Distance (MAD) \cite{Rusch2023ASO}. 
As seen in \cref{fig:oversmoothing}, all methods except for FSW-GNN and GIN exhibit oversmoothing starting from some $r$.

We note that long-range issues can be alleviated by graph rewiring methods, which effectively reduce problem radii \cite{gutteridge2023drew}, or by adding global information using spectral filters \cite{best_func_model} or graph transformers. A key distinction of FSW-GNN is that it performs well \emph{without modifying} the original graph topology. Thus, this work directly addresses the problem of relaying long-range messages, rather than circumventing it by architectural modifications.

\nd{Thinks to address about figs, if possible: (a) Table 2 text is too small. (b) Figure 3 I think tal was correct that we should send one half of it to an appendix, and discuss there why we get an essentially identical figure. (I think I understand why this could make sense)}

\section{Conclusion}
\label{sec_conclusion}
In this paper, we introduced FSW-GNN, the first bi-Lipschitz MPNN. Empirically, we found that FSW-GNN is highly effective, particularly for long-range problems, and appears to mitigate oversmoothing and oversquashing due to its inherent ability to preserve graph metrics.

A slight drawback of FSW-GNN is that its runtime is somewhat higher than that of standard MPNNs; see \cref{comparison_time_fswgnn}. An interesting direction for future work is to obtain quantitative estimates of its bi-Lipschitz constants.


\newpage

%
%

\clearpage


\PrintTheBibliography


\newpage
\appendix
\onecolumn

\begin{table}[H]
    \caption{Performance of different models on the LRGB results.}
    \label{tab:lrgb_datasets}
    \centering
    \begin{tabular}{lcc}
        \toprule
        Dataset & peptides-func (AP$\uparrow$) & peptides-struct (MAE$\downarrow$) \\ \hline
        GINE & 0.6621$\pm$0.0067 & \underline{0.2473$\pm$0.0017} \\
        GCN & 0.6860$\pm$0.0050 & \textbf{0.2460$\pm$0.0007} \\
        GatedGCN & 0.6765$\pm$0.0047 & 0.2477$\pm$0.0009 \\
        SortMPNN & \textbf{0.6914$\pm$0.0056} & 0.2494$\pm$0.0021 \\
        FSW-GNN  & \underline{0.6864$\pm$0.0048} & 0.2489$\pm$0.00155 \\ \hline
        SOTA & \textbf{0.73} & \textbf{0.242} \\ \hline
    \end{tabular}
\end{table}

\begin{table}[H]
\caption{Performance of different models on the MUTAG and Protein datasets.}
\label{tab:mutag}
\centering
\footnotesize
\setlength{\tabcolsep}{5pt} 
\renewcommand{\arraystretch}{0.95} 
\begin{tabular}{@{}lll@{}}
\toprule
\textbf{Model} & \textbf{MUTAG} & \textbf{Protein} \\
\midrule
GIN {\scriptsize\citep{gin}} & $89.4 \pm 5.6$ & $76.2 \pm 2.8$ \\
GCN & $85.6 \pm 5.8$ & $76.0 \pm 3.2$ \\
GraphSage & $85.1 \pm 7.6$ & $75.9 \pm 3.2$ \\
SortMPNN & $\mathbf{90.99 \pm 6.2}$ & \underline{$76.46 \pm 3.68$} \\
AdaptMPNN & $90.41 \pm 6.1$ & $75.12 \pm 3.64$ \\
FSW-GNN & \underline{$90.55 \pm 6.1$} & $\mathbf{76.93 \pm 7.64}$ \\
\bottomrule
SOTA & \underline{$96.66 \pm 1.23$} & $\mathbf{84.91 \pm 1.62}$ \\
\bottomrule
\end{tabular}
\end{table}
\section{Statistics on our benchmarks}
We present some statistics of the learning problems considered in the main text. In Table \ref{tab:common_graph_stats}, we present the statistics for each transductive dataset, and for the MUTAG dataset, including the number of nodes, edges, features, classes, average degree, and density, measuring the number of edges divided by the number of maximal edges. As we can see in the table, these datasets are very sparse.

Table \ref{tab:peptides_graph_stats} shows the same statistics for  the Peptides tasks from the  LRGB dataset.

\begin{table}[h!]
\caption{Graph statistics for transductive learning and MUTAG.}
\label{tab:common_graph_stats}
\centering
\resizebox{\textwidth}{!}{%
\begin{tabular}{@{}lcccccccccc@{}}
\toprule
\textbf{Dataset}     & \textbf{Cora} & \textbf{Cite.} & \textbf{Pubm.} & \textbf{Cham.} & \textbf{Squi.} & \textbf{Actor} & \textbf{Corn.} & \textbf{Texa.} & \textbf{Wisc.} & \textbf{MUTAG} \\ \midrule
\# Nodes             & 2708          & 3327           & 19717          & 2277           & 5201           & 7600           & 183            & 183            & 251            & 188            \\
\# Edges             & 5429          & 4732           & 44338          & 36101          & 217073         & 33544          & 295            & 309            & 499            & 744            \\
\# Features          & 1433          & 3703           & 500            & 2325           & 2089           & 931            & 1703           & 1703           & 1703           & 7              \\
\# Classes           & 7             & 6              & 3              & 5              & 5              & 5              & 5              & 5              & 5              & 2              \\
\textbf{Avg. Degree} & 4             & 2              & 4              & 31             & 83             & 8              & 3              & 3              & 4              & 8              \\
\textbf{Density}     & 0.0007        & 0.0009         & 0.0001         & 0.0159         & 0.0161         & 0.0012         & 0.0177         & 0.0184         & 0.0041         & 0.0422         \\ \bottomrule
\end{tabular}
}
\end{table}

\begin{table}[h!]
\caption{Graph statistics for Peptides datasets.}
\label{tab:peptides_graph_stats}
\centering
\begin{tabular}{@{}lcc@{}}
\toprule
\textbf{Dataset}       & \textbf{Peptides-Func} & \textbf{Peptides-Struct} \\ \midrule
\# Graphs              & 15,535                & 15,535                  \\
\# Nodes (Avg.)        & 150.94                & 150.94                  \\
\# Edges (Avg.)        & 2.04                  & 2.04                    \\
\textbf{Avg. Degree}   & 2.04                  & 2.04                    \\
\textbf{Density}       & \(1.74 \times 10^{-6}\) & \(1.74 \times 10^{-6}\) \\
\# Classes             & -                     & 10                      \\ \bottomrule
\end{tabular}
\end{table}
\section{Implantation details}
For the Transductive learning and the LRGB dataset, we looked for each learning rate in the set $1e^{-3},5e^{-3}$ and weights decay in the set $0.0,1e^{-5}$ and learning rate factor of 0.7. For each such configuration we stop according to the validation and take the best configuration of the validation set. For all Over-squashing tasks we took a learning rate of $0.001$ and weight decay $0$, and learning rate factor of 0.7. In all datasets we took the original dataset splits to train, test and validation sets.

\section{Relation to Tree Mover's Distance}\label{app:TMD}
\begin{figure*}[ht]
    \centering
    \begin{minipage}{0.6\linewidth}
        \centering
        \definecolor{darkgreen}{rgb}{0.0, 0.5, 0.0} 

\begin{tikzpicture}
\begin{axis}[
    width=\linewidth, 
    height=0.85\linewidth, 
    xlabel={MPNN iterations},
    ylabel={Distortion},
    xlabel style={font=\small},
    ylabel style={font=\small},
    legend style={
        at={(0.82,0.15)}, 
        anchor=south, 
        font=\small,
        nodes={scale=1, transform shape}, 
        legend columns=1 
    },
    xmin=0.5, xmax=9,
    ymin=0.1, ymax=1000000000000000,
    grid=both,
    grid style={opacity=0.5},
    ymode=log, 
    log basis y=10,
    xtick=data,
    ytick={1, 100, 10000, 1000000, 100000000, 10000000000, 1000000000000, 100000000000000},
    yticklabels={$1$, $10^2$, $10^4$, $10^6$, $10^8$, $10^{10}$, $10^{12}$, $10^{14}$},
    tick label style={/pgf/number format/fixed}
]

\addplot[
    color=blue, 
    mark=*, 
    mark size=2pt, 
    line width=1pt, 
    solid 
] table[x=Layers, y=GCN, col sep=comma] {empirical_distortion_vs_tmd.csv};
\addlegendentry{GCN}

\addplot[
    color=red, 
    mark=triangle*, 
    mark size=3pt, 
    line width=1pt, 
    dashed 
] table[x=Layers, y=GIN, col sep=comma] {empirical_distortion_vs_tmd.csv};
\addlegendentry{GIN}

\addplot[
    color=purple, 
    mark=diamond*, 
    mark size=3pt, 
    line width=1pt, 
    dotted 
] table[x=Layers, y=GAT, col sep=comma] {empirical_distortion_vs_tmd.csv};
\addlegendentry{GAT}

\addplot[
    color=orange, 
    mark=square*, 
    mark size=2pt, 
    line width=1pt, 
    dash dot 
] table[x=Layers, y=Sort-MPNN, col sep=comma] {empirical_distortion_vs_tmd.csv};
\addlegendentry{Sort-MPNN}

\addplot[
    color=darkgreen, 
    mark=square*, 
    mark size=2pt, 
    line width=1pt, 
    solid 
] table[x=Layers, y=FSW-GNN, col sep=comma] {empirical_distortion_vs_tmd.csv};
\addlegendentry{FSW-GNN}

\end{axis}
\end{tikzpicture}
    \end{minipage}
    \caption{Empirical distortion evaluation with respect to the Tree Mover's Distance (TMD)}
    \label{fig_empirical_distortion_tmd}
\end{figure*}

In this appendix we give a full description of 
the Tree Mover's Distance
as defined in \citet{TMD}.

We first review Wasserstein distances. Recall that if $(X,d) $ is a metric space, $\Omega \subseteq X$ is a subset,  then the Wasserstein distance can be defined on the space  of multisets consisting of $n$ elements in $\Omega$ via
\[
{W}_{d}( x_1,\ldots,x_n, y_1,\ldots,y_n) =\min_{\tau \in S_n} \sum_{j=1}^n d(x_j,y_{\tau(j)})
\]
For multisets of different size, the authors of \citep{TMD} used an augmentation map, which, for a fixed parameter $n$, augments multisets of size $r \leq n$ by padding with $n-r$ instances of an element $z$ in $X\setminus \Omega $, namely
$$\Gamma^z\left( x_1,\ldots,x_r \right)= (x_1,\ldots,x_r,x_{r+1}=z,\ldots.x_n=z)$$
and the augmented distance on multi-sets of size up to $n$ is defined by
$$W_{d}(X,\hat X)=W_{d}^0\br{\Gamma^z\of X, \Gamma^z\of {\hat X } }.$$ 

We now return to define the TMD. We consider the space of graphs $\GNof\Omega$, consisting of graphs with $\leq N$ nodes, with node features coming from a compact domain $\Omega \subseteq \RR^d$ such that $0\notin\Omega$. The TMD is defined using the notion of computation trees:
\begin{definition}
    (Computation Trees). Given a graph $\G=(V,E,X)$ with node features $\brc{ x_v }_{v\in V} $, let $\tree_v^1$ be the rooted tree with a single node $v$, which is also the root of the tree, and node features $x_v$. For $t \in \NN$  let $\tree_v^t$ be the depth-$t$ computation tree of node $v$ constructed by connecting the neighbors of the leaf nodes of $\tree_v^{t-1}$ to the tree. Each node is assigned the same node feature it had in the original graph $\G$. The multiset of depth-$t$ computation trees defined by $\G$ is denoted by $\mathcal{T}_G^K:=\brc{ \tree_v^t }_{v\in V}$.
    Additionally, for a tree $\tree$ with root $r$, we denote by $\mathcal{T}_r$ the multiset of subtrees that root at the descendants of $r$. 
\end{definition}

\begin{definition}
    (Blank Tree). A blank tree $\blankT$ is a tree (graph) that contains a single node and no edge, where the node feature is the zero vector $0$. 
\end{definition}

Recall that by assumption, all node features will come from the compact set $\Omega$, and $0 \not \in \Omega$. Therefore, the blank tree is not included in the space of trees with features in $\Omega$, and can be used for augmentation.

We can now define the tree distance:
\begin{definition}
    (Tree Distance).\footnote{This definition slightly varies from from the original definition in \cite{TMD}, due to our choice to set the depth weight to 1 and using the 1-Wasserstein which is equivalent to optimal transport.} 
    The distance between two trees $\tree_a, \tree_b$ with features from $\Omega$ and $0\not \in \Omega$, is defined recursively as
    \[
    \TD(\tree_a, \tree_b) := \begin{cases}
      \|x_{r_a}-x_{r_b}\|_1 + W_{\TD}^{\blankT} (\mathcal{T}_{r_a}, \mathcal{T}_{r_b}) & \text{if $t>1$} \\
      \|x_{r_a}-x_{r_b}\|_1 & \text{otherwise}  
    \end{cases}
    \]
    where $t$ denotes the maximal depth of the trees $\tree_a$ and $\tree_b$. Here $W_{\TD}^{\blankT} $ denotes the Wasserstein metric obtained from the metric $\TD$ on the space of trees of smaller depth, with augmentation by blank trees $\blankT$.
\end{definition}

\begin{definition}
    (Tree Mover's Distance). Given two graphs, $G_a, G_b$ and $w,t \ge 0$, the tree mover's distance is defined by
    \[
    \tmd^t(G_a, G_b) = W^{\blankT}_{\TD}(\mathcal{T}_{G_a}^t, \mathcal{T}_{G_b}^t),
    \]
\end{definition}
where $\mathcal{T}_{G_a}^t$ and $\mathcal{T}_{G_b}^t$ denote the multiset of all depth $t$ computational trees arising from the graphs $G_a$ and $G_b$, respectively. 
\citet{TMD} proved that $\tmd^t(G_a, G_b)$ is a pseudo-metric that fails to distinguish only graphs that cannot be separated by $t+1$ iterations of the WL test. Thus, assuming that $0\notin\Omega$, $\tmd^t(G_a, G_b)$ is WL equivalent on $\GNof\Omega$.

In addition, it is easy to see from the definition of TMD that it satisfies the following properties:
\begin{enumerate}
    \item  $\tmd^t((\A, \alpha \cdot \X),(\B, \alpha \cdot \Y)) = \alpha \cdot \tmd^t(\X,\Y)$ for any $\alpha \geq 0$.
    \item For fixed $\A,\B$, the TMD metric is piecewise linear in $\br{\X,\Y}$.
\end{enumerate}

These properties will be used to show that under the above assumptions, the embedding computed by FSW-GNN is bi-Lipschitz with respect to TMD.

\section{Proofs}
\label{app_proofs}
\subsection{DS metric}
\label{subapp_proofs_ds_metric}\label{metric_property}
Here we prove \cref{thm_rhods_is_wl_metric} 

\thmRhoDSIsWLMetric*

\begin{proof}
$\rhods$ is  symmetric,  since if a pair's minimum is obtained at $S$ then the exact value is obtained for the opposite pair with $S^{T}$ and vice-versa.

To prove the triangle inequality, let
Let $(V_1, \A,\X),(V_2, \B,\Y),(V_3, \C,\Z)$ three arbitrary graphs with $\abs{V_i}=n_i$, $i=1,2,3$. Then the following holds:
\begin{align*}
   \alpha:= \rhods(\A,\B) = \abs{n_1-n_2} + min_{S\in \Pi\of{n_1,n_2}}||\A S-S\B||_1 +\sum_{i,j}S_{i,j}||\X_{i}-\Y_{j}||_1 \\
  \beta:= \rhods(\B,\C) = \abs{n_2-n_3} + min_{S\in \Pi\of{n_2,n_3}}||\B S-S\C||_1 +\sum_{i,j}S_{i,j}||\Y_{i}-\Z_{j}||_1\\
  \gamma:=\rhods(\A,\C) = \abs{n_1-n_3} + min_{S\in \Pi\of{n_1,n_3}}||\A S-S\C||_1 +\sum_{i,j}S_{i,j}||\X_{i}-\Z_{j}||_1\\
\end{align*}
We want to prove $\gamma \leq \beta+\alpha$. Let $S^{1}, S^{2}$ be the minimizers of the first two equations. 
Define $S^{3} = n_2\cdot S^{1}\cdot S^{2}$ and note that $S^3 \in \Pi\of{n_1,n_3}$ because, using the notation $1_d$ to the all one $d$-dimensional vector, we have
\begin{align*}
S^31_{n_3}&=n_2\cdot S^{1}\cdot S^{2}1_{n_3}=S^{1}1_{n_2}=\frac{1}{n_1}1_{n_1} \\
1_{n_1}^TS^3&=n_2\cdot 1_{n_1}^TS_1\cdot S_2=1_{n_2}^TS_2=\frac{1}{n_3}1_{n_3}.
\end{align*}
Next, we prove the following lemma (as usual, norms in this lemma are Frobenius norms)
\begin{lemma}
Let $S \in \Pi\of{n_1,n_2}$, then for all $B\in \RR^{n_2\times b}$ and $A\in \RR^{a\times n_1}$

$$\|SB\|_1\leq \frac{1}{n_2}\|B\|_1, \quad \|AS\|_1\leq \frac{1}{n_1}\|A\|_1 $$
\end{lemma}
\begin{proof}

\begin{align*}
 \|SB\|_1&=\sum_{i=1}^{n_1}\sum_{j=1}^{b} |(SB)_{ij}|\\
 &\leq  \sum_{i=1}^{n_1}\sum_{j=1}^{b} \sum_{k=1}^{n_2} S_{ik}|B_{kj}|\\
 &=\sum_{j=1}^{b} \sum_{k=1}^{n_2} |B_{kj}| \cdot    \sum_{i=1}^{n_1}S_{ik}\\
 &=\frac{1}{n_2}\|B\|_1
\end{align*}
A symmetric argument can be used to show that $\|AS\|_1\leq \frac{1}{n_1} \|A\|_1$
\end{proof}

Given the lemma, we now have
\begin{align*}
    \frac{1}{n_2}||\A S^{3}-S^{3}\C||_1&=||\A S^{1}S^{2}-S^{1}S^{2}\C||_1=||\A S^{1}S^{2}-S^{1}\B S^{2}+S^{1}\B S^{2}-S^{1}S^{2}\C||_1 \\
    &\leq 
    ||\A S^{1}S^{2}-S^{1}\B S^{2}||_1 + ||S^{1}\B S^{2}-S^{1}S^{2}\C||_1 = 
    ||(\A S^{1}-S^{1}\B)\cdot S^{2}||_1
    \\
    &+
    ||S^{1}\cdot(\B S^{2}-S^{2}\C)||_1
    \leq 
    \frac{1}{n_2} ||\A S^{1}-S^{1}\B||_1 +
    \frac{1}{n_2}\cdot||\B S^{2}-S^{2}\C||_1
\end{align*}
Next, we show the second part is also smaller:
\begin{align*}
    \frac{1}{n_2}\sum_{i,j}S^{3}_{i,j}\cdot|\X_{i}-\Y_{j}|_1 =& \sum_{i,j}\sum_{k} S^{1}_{i,k}\cdot S^{2}_{k,j} \cdot ||\X_{i}-\Z_{j}||_1
    \\ \leq& \sum_{i,j}\sum_{k} S^{1}_{i,k}\cdot S^{2}_{k,j} \cdot (||\X_{i}-\Z_{k}||_1+||\Z_{k}-\Y_{j}||_1)\\
    =& 
    \sum_{i,j}\sum^{n}_{k=1} S^{1}_{i,k} \cdot S^{2}_{k,j} \cdot ||\X_{i}-\Z_{k}||_1+\sum_{i,j}\sum^{n}_{k=1} S^{1}_{i,k} \cdot S^{2}_{k,j} \cdot ||\Y_{j}-\Z_{k}||_1 
\end{align*}
We now open both sums:
\begin{align*}
    \sum_{i,j}\sum_{k} S^{1}_{i,k} \cdot S^{2}_{k,j} \cdot ||\X_{i}-\Z_{k}||_1 =&\sum_{i}\sum_{k} S^{1}_{i,k}  \cdot ||\X_{i}-\Z_{k}||_1 \cdot \sum_{j}  S^{2}_{k,j}
    \\=&\frac{1}{n_2} \sum_{i}\sum_{k} S^{1}_{i,k}  \cdot ||\X_{i}-\Z_{k}||_1 
\end{align*}
With the same argument, we obtain that 
\begin{align*}
    \sum_{i,j}\sum_{k}S^{1}_{i,k}\cdot S^{2}_{k,j}\cdot ||\Y_{j} - \Z_{k}||_1 = \frac{1}{n_2} \sum_{i,j} S^{2}_{i,j}\cdot ||\Y_{i} - \Z_{j}||_1
\end{align*}
So overall we obtain
\begin{align*}
\alpha \leq& \abs{n_1-n_3} +||\A S^{3}-S^{3}\C||_1 +\sum_{i,j}S^{3}_{i,j}||\X_{i}-\Z_{j}||_1 \\
\leq &\abs{n_1-n_2}+ ||\A S^{1}-S^{1}\B||_1 +\sum_{i,j}S^{1}_{i,j}||\X_{i}-\Y_{j}||_1+\abs{n_2-n_3}
\\+&||\B S^{2}-S^{2}\C||_1 +\sum_{i,j}S^{3}_{i,j}||\Y_{i}-\Z_{j}||_1 
 = \alpha + \beta
\end{align*}

Now, we show that our metric $\rhods$ is equivalent to WL. Clearly, any pair of graphs with different numbers of vertices are distinguished both by WL and by $\rhods$. Thus, in the following, we assume that the two graphs have the same number of vertices.

We begin first with some needed definitions.
\paragraph{Partitions}
Most of our techniques are inspired by \cite{book}.
Given $\G = (\A, \X)$, we define a stable partition $\V = \P_{1}\cup \P_{2}...\cup \P_{k}$ if 
\begin{align}
    \forall u,v \in \P_{i}, \X_{u} =& \X_{v} \\
    \forall l \in [k],i \in [k], u, v \in  \P_{i}, |\N_{u}\cap \P_{l}| =& |\N_{v} \cap \P_{l}| 
\end{align}
In simple words, two nodes in the same partition must have the same feature and the same number of neighbors with the same feature.
Note that the union of all singleton is a valid stable partition.
We can characterize a partition as a $k$ tuple, and in the $i'th$ place, we put a tuple of the common feature and a vector telling the number of neighbors in other partitions.
We say that graphs have the same stable partition; if, up to permitting of the $k$ tuple indices, we have the same $k$ tuple. 
\begin{lemma}
    \label{stable}
    The number of colors in 1-WL can't increase at level $n$. In addition, all nodes with the same color at step $n$ make a stable partition.
\end{lemma}
\begin{proof}
The number of colors of $1$-WL can only increase as the number of iterations increases. It is also known that, if at some step $t$ the number of colors does not increase, then the process is 'stuck'. Accordingly,  after at most $n$ iterations, the number of different colors will stay the same.  Denote by $\P = (\P_{1},..,\P_{s})$ a partition of the nodes with the same coloring at the $n$ iteration, and be $c\in C$ some color, we claim that nodes in the same partition, have the same number of neighbors with color $c$. As otherwise, those nodes that now have the same coloring will have different coloring in the next $n+1$ iteration. But as the number of colors doesn't decrease (because of the concatenation of the current color), we have at least one more color, a contradiction. So, we found a stable partition.
\end{proof}
\begin{lemma}
    \label{equivalence}
    The following conditions are equivalent:
    \begin{itemize}
        \item $\G \cong_{1-WL} \HH$
        \item Both graphs have a common stable partition. That is, there is a relabeling of $G$ such that the same partition $\P_{1},..,\P_{s} $ of $V$ is a stable partition for both $\G$ and $\HH$, and the features inside each partition are the same, in both $\G$ and $\HH$.   
    \end{itemize}
\end{lemma}
\begin{proof}
    Assume we have the same common partition.
    Up to renaming the names of the vertices of the nodes in the first graphs, we assume we have the same stable partition with the same parameters.
    Assume that a common stable partition exists $\P = \P_{1}\cup..\cup P_{k}$.  
    Then, by simple induction on the number of iterations, we will prove that all nodes in the same partition have the same color in both graphs.
    \newline
    \textbf{Basis} 
    Nodes in the same partition have the same initial feature and, thus, have the same color (in both graphs).
    \newline
    \textbf{Step}
    By the induction hypothesis, all nodes in the same partition iteration $T$ have the same color; 
    and they both have the same number of neighbors with the same color, so the aggregation yields the same output.
    Thus, in iteration $T+1$, those nodes also have the same color.
    Note that this argument is symmetric to both graphs; thus, the 1-WL test will not distinguish them.
    \ys{Nadav take a look}
    On the other hand, if $\G_1,\G_2$ have the same 1-WL embedding, then we partition the nodes to those classes with the same color at iteration $n$, denoted by $\P = \P_{1}\cup..\cup P_{k}$. 
    We have to show this partition is valid.
    First, by definition, those nodes $u,v \in \P_{i}$ have the same node feature (by iteration 1); next, if $u,v$ don't have the same number of neighbors in $\P_{l}$ for some $l\in [k]$, then their color won't be the same at iteration $n+1$. But, as shown in lemma \ref{stable}, the number of distinct colors can't increase between $n$ to $n+1$. Thus, we found a common stable partition.            
\end{proof}
Before proving the following lemma, we revise a definition from \cite{book}.
We define direct the direct sum of matrices \( S_1 \) and \( S_2 \) by:
\[
S_1 \oplus S_2 =
\begin{bmatrix}
S_1 & 0 \\
0 & S_2
\end{bmatrix}
\]
Given $S \in \R^{n \times n}$, we say $S$ is composable if there exists $P,Q,S_1,S_2$ such that 
\begin{align*}
    S = P \cdot (S_{1} \oplus S_{2}) \cdot Q
\end{align*}
Such that $P,Q$ are permutation matrices.
By simple induction, we can write $M$ as a direct sum of an indecomposable
\begin{align*}
     S = P \cdot (S_{1} \oplus S_{2}\oplus ... \oplus S_{t}) \cdot Q
\end{align*}
\begin{lemma}
    \label{features}
    Let $S \in \Pi\of{n,n}$ and assume it's in the form of $S = S_{1} \oplus ... \oplus S_{k}$ such that all blocks are indecomposable and $\sum_{i,j}S_{i,j}\cdot||\X_{i}-\Y_{j}||=0$. Denote by $i_{1},...,i_{k+1}$ the indices define the start and the end of $S_{1},...,S_{t}$ and by $I_{k}:=[i_{k},i_{k+1}]$.
    Then $\X_{i}=\Y_{j},\forall t \in [k], \forall i,j \in I_{t}$.
\end{lemma}
\begin{proof}
    Given $t\in [k]$, we build a bipartite graph from $I_{t}$ to itself, such that two indexes $i,j$ are connected if $S_{i,j}>0$. Note that this graph is connected, as otherwise, $S_{t}$ could be composed into its connected components, and thus $S_{t}$ would be composable. 
    Given a path $P = (i_{1},..., i_{l})$ in the graph, note that by definition, as the metric vanishes, and $S_{i_{j}, i_{j+1}}>0$, so $\X_{i_{j}}=\Y_{i_{j+1}}$, thus $\X_{i}=\Y_{j},i\in I_{k},j \in I_{k}$ and we are done.
\end{proof}
\begin{theorem}
    Be $\G,\HH$ two featured graphs, then 
    \begin{align*}
        d(\G,\HH) = 0 \iff \G \wleq \HH
    \end{align*}
\end{theorem}
We first prove that if the two graphs are 1-WL equivalent, then this metric vanishes. We may assume they have the same stable partition as we proved above. Take $\P=\P_{1}\cup..\cup \P_{t}$, rename the nodes such that they come in consecutive order and denote by $n_{1},n_{2},..,n_{k}$ the sizes of the partitions. Note that by definition, $\forall u,v\in \P_{i}$, both have the same feature and number of neighbors in all $\P_{l}$.
As in the book \cite{book}, define $S:=\frac{1}{n_{1}}J_{n_{1}}\oplus...\oplus\frac{1}{n_{j}}J_{n_{k}}$ and from the book \cite{book}, we know that $\A S=S\B$. As all nodes in the same partition have the same feature, $\X_{i}=\Y_{j},\forall i,j \in I_{k} = [n_{t},n_{t+1}],\forall t \in [k]$, so as $S$ is non-zero only on indexes in the same partition then also this metric vanishes, so also the sum vanishes on $S$. 

For the next direction, be $\G = (\A,\X),\HH = (\B,\Y)$ two graphs and assume there exists a matrix of the form of $S=P(S^{1}\oplus...\oplus S_{t})Q$ such that the metric vanishes and denote by $D =S^{1}\oplus...\oplus S_{t}$
We show we can choose $S$ such that it's a block matrix.
\begin{align*}
    ||PDQ\cdot \A - \B \cdot  PDQ|| =& ||P\cdot (DQ\A\cdot Q^{-1}-P^{-1}\B P\cdot D )\cdot Q|| \\
    \stackrel{(*)}{=}& ||DQ\A\cdot Q^{-1}-P^{-1}\B P\cdot D||  \\
    =& \sum_{i,j} S_{i,j} \cdot |\X_{i} - \Y_{j}| \sum_{i,j} (PDQ)_{i,j} \cdot |\X_{i} - \Y_{j}|\\
    =& 
    \sum_{i,j} D_{\pi_{1}^{-1}(i),\pi_{2}(j)} \cdot |\X_{i} - \Y_{j}|  \\
    =&\sum_{i,j} D_{i,j} \cdot |\X_{\pi_{1}(i)} - \Y_{\pi_{2}^{-1}(j)}|,
\end{align*}
where (*) follows from the fact that permutation matrices preserve the norm.
We define two graphs $\hat{\G} = (Q\A Q^{-1},Q\X)\cong \G, \hat{\HH} = (P^{-1}\A P,P^{-1}\Y)\cong \HH$.
So, we can choose $S$ to be a diagonal block matrix.
Note that $D$ defines a partition $\P_1,\ldots,\P_s $ of the nodes, and we will prove that it's a stable partition of both graphs.
From $SA=BS$, we obtain, as in the book \cite{book}, that each of the two nodes $u,v\in \P_{k}$ has the same number of neighbors in $\P_{l},\forall l\in [s]$. 
Next, by the lemma \ref{features}, we know that $\X_{i} = \Y_{j},\forall i,j \in \P_{k}$, so nodes in the same partition have the same feature. Thus, this partition is stable.
So, $\hat{\G},\hat{\HH}$ have exactly the same partition.
Then $\G,\HH$ have the same partition up to isomorphism, and by lemma \ref{equivalence}, both graphs are 1-WL equivalent.
\end{proof}

\subsection{FSW-GNN: Equivalence to WL}
\label{subapp_proofs_fswgnn_wl_equivalence}
We now prove \cref{thm_fswgnn_is_wl_equivalent} that says the FSW-GNN is equivalent to WL.

\thmFSWGNNIsWLEquivalent*
\begin{proof}
This proof employs the finite witness theorem \cite{amir2023neural}, and the basic methodology to use it for graph neural network injectivity, first presented by \citet{hordan2024complete}. 

It is sufficient to show that the functions $\Phi^{\br{t}}$ and $\Psi$ are injective vector-to-vector functions, and that the FSW ebmedding maps used in the FSW-GNN are injective multiset to vector functions. The injectivity of all these functions ensures that the MPNN is WL-equivalent \cite{gin}.

By Theorem 4.1 in \citep{amir2024injective}, if we take  the FSW embedding with an output dimension of $2Nd+1$, defined on the space of multisets with at most $N$ features in $\RR^d$, then the embedding will be injective for Lebesgue almost every choice of its parameters. This injectivity is in the sense that two multisets will attain the same value if and only if they define the same distribution. This can happen when the multisets are not equal, but have different cardinalities. However, the last coordinae of the FSW embedding, which encodes the multiset cardinality, ensures that this will not be an issue. Accordingly the FSW embedding is injective on the multiset space. 

We note that the output of the FSW embedding will now be node features of a higher dimension of $2Nd+2$, rather than just $d$. However, the set of all possible features obtrained from the previous iteration will be a semialgebraic set of dimension $\leq Nd$. As long as  this is the case, reapplying the FSW embedding with output dimension of $2Nd+2$ will still be injective, for almost every choice of parameters.

The same argument can also be applied for the linear functions $\Phi^{\br{t}}$ and $\Psi$ used in FSW-GNN: It is an easy consequence of the finite witness theorem \cite{amir2023neural,dym2023low}, that almost every $(2Nd+1)\times m $ matrix, applied to a semialgebraic subset of $\RR^m$ of dimension $\leq Nd$, will be injective. This concludes the proof.
\end{proof}
\subsection{FSW-GNN: Bi-Lipschitzness}
\label{subapp_proofs_fswgnn_bilipschitz}
We first prove \cref{thm_pwl_ratio_bound}.
\thmPWLRatioBound*
\nd{olde proof commented out}
\begin{proof}
  
   By definition, there is a partition of $M$ into a finite union of compact polygons, so that the restriction of $f$ to each polygon is an affine function. $g$ also has such a partition, and by taking the mutual refinement of the two partitions, we obtain a finite partition of $M$ into compact polygons, such that the restriction of both $f$ and $g$ to each polygon is affine. 

   Accordingly, it is sufficient to prove the claim on a single compact polygon $P\subseteq M$ on which the restriction of both $f$ and $g$ is affine. A compact polygon has a finite number of extreme points $v_1,\ldots,v_k$, and every point in the polygon can be written as a convex combination of these points. Now let us denote 
   $$I=\{i\in [k]| \quad f(v_i)>0 \}=\{i\in [k]| \quad g(v_i)>0 \} .$$
 \nd{Yonatan for your question:} 
  We note that if $I$ is empty, then $f$ and $g$ are identically zero on $P$, and there is nothing to prove. Otherwise, the following quantities are well defined and strictly positive:
  
   \begin{align*}
   m_f&=\min_{i\in I} f(v_i), \quad M_f=\max_{i\in I} f(v_i)\\ 
   m_g&=\min_{i\in I} g(v_i), \quad M_g=\max_{i\in I} g(v_i)
\end{align*}
Then for every $x\in P$, there exist non-negative $\theta_1,\ldots,\theta_k $ which sum to $1$ such that $x=\sum_{i=1}^k \theta_iv_i$. 

It follows that 
\begin{align*}
f(x)&=f(\sum_{i=1}^k \theta_iv_i)
=\sum_{i=1}^k \theta_i f(v_i)
=\sum_{i\in I} \theta_i f(v_i)\\
&\geq m_f \sum_{i\in I} \theta_i
=\frac{m_f}{M_g}\cdot M_g \sum_{i\in I} \theta_i
\geq \frac{m_f}{M_g}\cdot \sum_{i\in I} \theta_i g(v_i)\\
&= \frac{m_f}{M_g}\cdot g(\sum_{i=1}^k \theta_i v_i)
=\frac{m_f}{M_g}\cdot g(x).
\end{align*}    
By reversing the roles of $f$ and $g$, we obtain that for every $x\in P$,
$$g(x) \geq \frac{m_g}{M_f}f(x) .$$
This concludes the proof.
   \end{proof}

We now turn to the full proof of \cref{thm_fswgnn_is_bilip}.

\thmFSWGNNIsBilip*

\begin{proof}
Following the reasoning in the main text, the rest of the proof is as follows: Let $G,\tilde G \in \GNof\Omega$, $G=\br{V,E,X}$ and $\tilde G=\br{\tilde V, \tilde E, \tilde X}$, with $n$ and $\tilde n$ vertices respectively. Since there is a finite number of choices of possible $n,\tilde n \leq N$ and edges $E$, $\tilde E$ It is enough to show that once these parameters are chosen, the bi-Lipschitz ratio is bounded on all choices of $X \in \Omega^n$, $\tilde X \in \Omega^{\tilde n}$, . Define the function $f\of{X,\tilde X} = \norm{\bh_G - \bh_{\tilde G}}_1$, $g\of{X,\tilde X} = \rho\of{G,\tilde G}$, where $\rho$ is either $\rhods$ or TMD. By the comment above, both $f$ and $g$ are piecewise linear. Since both FSW-GNN and the metric $\rho$ are WL-equivalent, $f$ and $g$ have the same zero set. Lastly, since $\Omega$ is compact and does not contain $\bzero$, the distance from $\Omega$ to $\bzero$ is strictly positive, and thus $\Omega$ can be extended to a compact polygon that does not contain $\bzero$ (e.g. by taking the set-difference between a compact box that contains $\Omega$ and a small $\ell_{\infty}$ ball around $\bzero$ that is disjoint with $\Omega$). Thus, \cref{thm_pwl_ratio_bound} guarantees the existence of Lipschitz constants $c,C$.

Finally, note that by equivalence of all norms on a finite dimensional space, the same bi-Lipschitz equivalence holds if we replace the one-norm in the definition of $f$ with any other norm.
Q.E.D.
\end{proof}

\thmNodeLevelBiLip
\begin{proof}
Let $G = \br{V,A,\bX}$, $\tilde G = \br{\tilde V, \tilde A, \btX}$ be two graphs in $\GNof\Omega$. Let $t\in\NN$ denote the number of MPNN/WL iterations iterations applied. Let $v\in V$, $\tilde v \in \tilde V$. Let $\bh^{\br{t}}_v$ and $\bh^{\br{t}}_{\tilde v}$ be the output features of vertices $v$ and $\tilde v$ computed by FSW-GNN. Then $\textup{TD}\of{\tree_v^{(t)},\tree_{\tilde v}^{(t)}} = 0$ if and only if $v$ and $\tilde v$ are assigned the same features by the WL test, which in turn takes place if and only if $\bh_v^{(t)} = \bh_{\tilde v}^{(t)}$. Note that $\norm{\bh_v - \bh_{\tilde v}}_1$ is a piecewise-linear function of the input vertex features $\brc{\bx_v}_{v \in V}$ and $\brc{\btx_{\tilde v}}_{\tilde v \in \tilde V}$. Moreover, $\textup{TD}\of{\tree_v^{(t)},\tree_{\tilde v}^{(t)}} = 0$ can be transformed with bounded distortion to a piecewise-linear function of $\brc{\bx_v}_{v \in V}$ and $\brc{\btx_{\tilde v}}_{\tilde v \in \tilde V}$. Lastly, similarly to the proof of \cref{thm_fswgnn_is_bilip}, $\Omega$ can be extended to a compact polygon that does not contain $\bzero$. Thus, by \cref{thm_pwl_ratio_bound}, there exist constants $0<c,C < \infty$ such that
\begin{equation*}
c \cdot\textup{TD}\of{\tree_v^{(t)},\tree_{\tilde v}^{(t)}}
\leq
\norm{\bh_v^{(t)} - \bh_{\tilde v}^{(t)}}_1
\leq
C \cdot \textup{TD}\of{\tree_v^{(t)},\tree_{\tilde v}^{(t)}}.
\end{equation*}
\end{proof}

\section{Experiment Details}
\label{app_numerical_experiments}
We used the Adam optimizer for all experiments.

For the empirical distortion evaluation, we used pairs of graphs $G$, $\tilde G$, each of which consisting of four vertices and the edges $1-2-3-4-1$. Two random vectors $v_0, \Delta v \in \RR^d$ were drawn i.i.d. Gaussian and normalized to unit length. In $G$ all vertex features were set to $v_0$, whereas in $\tilde G$ they were set to $v_0 + \varepsilon \sigma \Delta v$, with $\sigma = 1$ for $v_2, v_4$ and $-1$ for $v_1, v_3$. We used $\varepsilon = \brc{1,10^{-1},10^{-2},\ldots,10^{-6}}$, and generated 100 pairs for each value of $\varepsilon$, and evaluated the constants $C$, $c$ for the resulting pairs. This experiment was repeated 10 times and the average distortion was taken. To ensure accurate results, we used 64-bit floating-point arithmetic in this experiment.

For the NeighborsMatch problem from \citep{alon}, we used the protocol developed in their paper: we used  their implementation for the MPNNs we compared to, with a hidden dimension of $64$ for all models, searched for each of its best hyper-parameters, and reported the training accuracy. For fair comparison with rival models, we repeated each sample 100 times, as was done in \cite{alon}.

For the Ring dataset, we used the results from \cite{bron} and trained our models with a hidden dimension of 64.

For the LRGB dataset, we trained all models under the constraint of 500K parameters. In contrast, for the MolHIV dataset, there was no restriction, and we trained the models with 40K parameters.

For the transductive learning tasks, we used a hidden dimension of 128 across all models.

\paragraph{Timing} Here, we show the testing time for FSW-GNN, GIN, and GCN, GGNN, GAT, and GraphSAGE on the MUTAG graph classification task.
\begin{table}[h!]
    \caption{Comparison of Average Time in seconds per testing in evaluation mode for different models on MUTAG.}
    \label{comparison_time_fswgnn}
    \centering
    \begin{tabular}{|c|c|}
        \hline
        Model          & Avg Time per Testing  \\ \hline
        GIN            & 0.09                    \\ \hline
        GCN            & 0.08                    \\ \hline
        GAT            & 0.09                    \\ \hline
        GGNN           & 0.08                    \\ \hline
        GraphSage      & 0.07                    \\ \hline
        FSW-GNN        & 0.11                 \\ \hline
    \end{tabular}
\end{table}

\section{Ordered subgraph aggregation}\label{app:subgraph}
In this section, we recall sub-graph aggregation networks and show that FSW-GNN could be used to create a Bi-lipshiz embedding with respect to the metric equivalent to sub-graph WL.
\subsection{Ordered subgraph 1-WL}. 
The OSWL (Ordered sub-graph WL) is taken from \cite{OSWL,OSWL2}.
Given 1-WL procedure using $T$ iterations denoted by $WL^{T}$, and deterministic graph sampling $\Pi$(could be node/edge deletion), we define the following three steps: 
\begin{itemize}
    \item Step 1: Deterministic graph sampling (e.g. node/edge deletion) $\Pi(\G) = \{\G_1,\ldots,\G_m\}$ .
    \item $y_i = WL^{T}(\G_i),\G_i \in \Pi(\G)$
    \item $y_G = \phi(\{y_1,\ldots,y_m\})$
\end{itemize}
Where $WL^T $ denotes the application of $T$ iterations of the WL test, and $\phi$ is an injective multiset function.
\begin{definition}
    We say $\G_1, \G_2$ to be equivalent up to OSWL, if $OSWL^{T}(\G_1) = OSWL^{T}(\G_2),\forall T \in \NN$, and denoted by $\G_1 \cong \G_2$. As for WL, checking equality for $T =|\G_1|$ is sufficient to ensure equality for all $T$.
\end{definition}
\paragraph{Ordered subgraph MPNN}
Ordered subgraph MPNNs are very similar to OSWL. Given an MPNN and a multi-set to vector embedding $\phi$, we define its sub-graph version: 
\begin{itemize}
    \item Step 1: Deterministic graph sampling  $\Pi(\G) = \{\G_1,\ldots,\G_m\}$ 
    \item $y_i = MPNN(\G_i),\G_i\in \Pi(\G)$
    \item $y_G = \phi(\{y_1,\ldots,y_m\})$
\end{itemize}
\paragraph{OSWL equivalent metric}
We define a metric OSWL equivalent similarly to \ref{WL_metric}:
\begin{definition*}[WL metric]
    A \emph{OSWL metric on $\GNofe\Omega$} is a function $\rho:\GNofe\Omega \times \GNofe\Omega \to \RR_{\geq 0}$ that satisfies the following conditions for all $G_1, G_2, G_3 \in \GNofe\Omega$:
    \begin{subequations}
    \begin{align}        
        \rho\of{G_1, G_2} &\eqx \rho\of{G_2, G_1}
        & \quad &\textup{\emph{Symmetry} }
        \\ \rho\of{G_1, G_3} &\leqx \rho\of{G_1, G_2} + \rho\of{G_2, G_3}
        & \quad &\textup{\emph{Triangle inequality} }
        \\ \rho\of{G_1, G_2} &= 0 \Longleftrightarrow G_1 \cong G_2.
        & \quad &\textup{\emph{OSWL equivalence} }
    \end{align}
    \end{subequations}
\end{definition*}
\begin{theorem}
    Be a 1-WL equivalent metric $d_{WL}$, and sampling procedure, the following metric is OSWL equivalent:  
    \begin{align*}
    d_{sub}(\G,\HH) = min_{\pi \in S_m}\sum^{m}_{i=1} d_{WL}(\G_i,\HH_{\pi(i)})
\end{align*}
\end{theorem}
\begin{proof}
    We prove only the last part, which is not trivial.
    Assume first that the number of nodes is the same, then:
    \begin{align*}
        d_{sub}(\G,\HH) = 0 \iff \exists \pi \in S_m:\forall i\in [m]: d_{WL}(\G_i,\HH_{\pi(i)}) = 0 \iff \\ 
        \iff \exists \pi \in S_m: \forall i \in [m],y^{\G}_i = y^{\HH}_{\pi(i)}  \\
        \phi(\{y^{\G}_1,\ldots,y^{\G}_m\}) = \phi(\{y^{\HH}_1,\ldots,y^{\HH}_m\}) \iff OSWL(\G) = OSWL(\HH)
    \end{align*}
\end{proof}
\begin{theorem}
    Assume we use FSW-GNN as MPNN, FSW aggregation as $\phi$, $d_{DS}$, as the 1-WL equivalent metric, then the overall OSMPNN  is OSWL bilipshiz.
\end{theorem}
\begin{proof}
    Note that the OSMPNN is piece-wise linear as a composition of piece-wise linear functions. Also, $d_{sub}$ is piece-wise linear a summation of piece-wise linear functions. In addition, both functions vanish on the same set, so by the \ref{thm_pwl_ratio_bound}, the OSMPNN is bilipshiz.
\end{proof}


\end{document}